\documentclass[10pt,twocolumn,letterpaper]{article}

\usepackage[final]{cvpr}              

\usepackage{graphicx}
\usepackage{amsmath}
\usepackage{amssymb}
\usepackage{booktabs}

\definecolor{cvprblue}{rgb}{0.21,0.49,0.74}
\usepackage[pagebackref,breaklinks,colorlinks,allcolors=cvprblue]{hyperref}

\usepackage[utf8]{inputenc} 
\usepackage[T1]{fontenc}    
\usepackage{hyperref}       
\usepackage{url}            
\usepackage{booktabs}       
\usepackage{amsfonts}       
\usepackage{xspace} 
\usepackage{nicefrac}       
\usepackage{microtype}      
\usepackage{xcolor}         
\usepackage{algorithm}
\usepackage{algorithmic}
\usepackage{mathtools}
\usepackage{empheq}

\usepackage{multirow}
\usepackage{array}
\usepackage{siunitx}
\usepackage{adjustbox}
\usepackage{subcaption}
\usepackage{graphicx}
\usepackage{wrapfig}
\usepackage{comment}
\usepackage{amssymb}
\usepackage{enumitem}

\usepackage{cuted} 
\usepackage{etoolbox} 
\usepackage{multicol}

\usepackage{amsmath}
\usepackage{amssymb}
\usepackage{amsthm}

\newtheorem{lemma}{Lemma}
\newtheorem{theorem}{Theorem}
\newtheorem{remark}{Remark}

\usepackage{stackrel}

\newcommand{\ra}[1]{\renewcommand{\arraystretch}{#1}}

\newcommand{\Method}{DDiff\xspace}
\newcommand{\method}{DDiff\xspace}

\def\paperID{} 
\def\confName{CVPR}
\def\confYear{2026}

\title{Dual Ascent Diffusion for Inverse Problems}

\author{Minseo Kim \qquad Axel Levy \qquad Gordon Wetzstein\\
Stanford University\\
{\tt\small \{kminseo, axlevy, gordonwz\}@stanford.edu}
}

\begin{document}
\maketitle
\begin{abstract}
Ill-posed inverse problems are fundamental in many domains, ranging from astrophysics to medical imaging. Emerging diffusion models provide a powerful prior for solving these problems. Existing maximum-a-posteriori (MAP) or posterior sampling approaches, however, rely on different computational approximations, leading to inaccurate or suboptimal samples. To address this issue, we introduce a new approach to solving MAP problems with diffusion model priors using a dual ascent optimization framework. Our framework achieves better image quality as measured by various metrics for image restoration problems, it is more robust to high levels of measurement noise, it is faster, and it estimates solutions that represent the observations more faithfully than the state of the art.
\end{abstract}    
\section{Introduction}
\label{sec:intro}

We are interested in solving inverse problems, where an unknown image or signal $\mathbf{x}$ is estimated from noisy and corrupted observations $\mathbf{y}$. 
These types of problems arise across science and engineering, for example, in image restoration~\cite{liang2021swinir}, astrophysics~\cite{akiyama2019first}, medical imaging~\cite{song2021solving}, protein structure determination~\cite{fadini2025alphafold, maddipatla2024generative, levy2024solving}, among other domains.
In all cases, a linear or nonlinear function $\mathcal{A} \left( \cdot \right)$ models a domain-specific image formation process. Although the likelihood of observations $p \left( \mathbf{y} | \mathbf{x} \right)$ depends on the statistical model of the noise in the observations, closed-form expressions exist for specific cases. 
For example, the image formation model for zero-mean Gaussian i.i.d.\ noise with variance $\sigma^2$ is $\mathbf{y} = \mathcal{A} \left( \mathbf{x} \right) + \mathcal{N} \left( \mathbf{0}, \sigma^2 \right)$ and its log-likelihood is $\log p \left( \mathbf{y} | \mathbf{x} \right) = \frac{-1}{2 \sigma^2} \left\| \mathbf{y} - \mathcal{A} \left( \mathbf{x} \right) \right\|_2^2$, up to an additive term that does not depend on $\mathbf{x}$. 
Given the likelihood of the observations and a prior $p \left( \mathbf{x} \right)$, inverse problem solvers aim at either \emph{maximizing} or \emph{sampling from} the posterior $p(\mathbf{x}|\mathbf{y})\propto p(\mathbf{y}|\mathbf{x})p(\mathbf{x})$ in a Bayesian framework. 
Most inverse problems are ill-posed, making the prior a crucial component of the solution-finding process.

Maximum-a-posteriori (MAP) approaches aim at maximizing the posterior to find the most likely solution given a prior. Early approaches used ``hand-crafted'' priors to promote smoothness,  piece-wise constancy via Total Variation~\cite{rudin1992nonlinear, beck2009fast}, or sparsity in a transform domain~\cite{donoho2006compressed}, while most modern approaches use some form of neural network~\cite{zhang2017beyond}. The plug-n-play (PnP) approach~\cite{pnp}, for example implemented by the Alternating Direction Method of Multipliers (ADMM)~\cite{admm} algorithm, is a popular and versatile framework to solve MAP problems by leveraging (Gaussian) denoisers as priors. MAP finds the single, most likely solution to an inverse problem. However, oftentimes one is interested in sampling from the posterior of all feasible solutions. For this reason, many recent works~\cite{score-ald, score-sde, ilvr, dps, pigdm, snips, ddrm, ddnm, score-prior, red-diff, daps} focus on posterior sampling using powerful pretrained diffusion models as priors, as surveyed in~\citep{survey}. While these recent diffusion posterior sampling methods show great promise, they are all fundamentally limited by the optimization framework that is used to combine the likelihood of the image formation model and the prior during optimization.

In this work, we do not aim to develop a method that provably samples from the posterior, but instead
focus on deriving an optimization strategy that accurately and efficiently solves the MAP problem
using a prior given by a pretrained diffusion model and a dual ascent–based optimization framework
inspired by ADMM~\cite{admm}. Our approach, dubbed \emph{\Method}, is faster and shown to achieve better reconstruction quality compared with the state of the art for image restoration problems, including single-image super resolution, inpainting, deblurring, phase retrieval, and high-dynamic range imaging. Moreover, \Method is more robust to high levels of measurement noise, and our reconstructions more faithfully model the observations by exhibiting closer-to-zero residuals than existing methods. The latter is important because the log-likelihood of the observations $\textrm{log} \, p \left( \mathbf{y} | \mathbf{x} \right)$ is an indicator for the level of hallucination a generative prior, such as a diffusion model, introduces when computing a solution. Notably, we establish the first fixed-point convergence proof for a diffusion-based posterior optimization method and extend our framework to a latent setting for efficient inference in compact feature spaces.


\section{Background on Inverse Problems}
\subsection{Maximum-a-Posteriori Solutions}
\label{sec:map}


A maximum-a-posteriori (MAP) solution aims to find the solution $\mathbf{x}_{\textrm{MAP}}$ that maximizes the posterior $p \left( \mathbf{x} | \mathbf{y} \right) \propto p \left( \mathbf{y} | \mathbf{x} \right) p \left( \mathbf{x} \right)$. Typically, this is done by minimizing the negative log-likelihood as 
\begin{equation}
\begin{aligned}
\mathbf{x}_{\mathrm{MAP}}
&= \operatorname*{argmin}_{\mathbf{x}} \; -\big( \log p(\mathbf{y}\mid \mathbf{x}) + \log p(\mathbf{x}) \big) \\
&= \frac{1}{2\sigma^2} \left\| \mathbf{y} - \mathcal{A}(\mathbf{x}) \right\|_2^2 \;-\; \log p(\mathbf{x})
\end{aligned}
\label{eq:map}
\end{equation}

%

The alternating direction method of multipliers (ADMM)~\cite{admm} is a common approach to solving the MAP problem. ADMM attempts to blend the benefits of dual decomposition and augmented Lagrangian methods for constrained optimization. 
For this purpose, a slack variable $\mathbf{z}$ is introduced to split the objective function in Eq.~\ref{eq:map} into a data fidelity term $- \log  p \left( \mathbf{y} | \mathbf{x} \right)$ and the log-prior term $- \log  p \left( \mathbf{z} \right)$, subject to $\mathbf{x} = \mathbf{z}$. 
%
ADMM then forms the Augmented Lagrangian of the split formulation as
%
\begin{equation}
\begin{aligned}
    L_\rho \left( \mathbf{x}, \mathbf{z}, \mathbf{u} \right) &= \frac{1}{2 \sigma^2} \left\| \mathbf{y} - \mathcal{A} \left( \mathbf{x} \right) \right\|_2^2 - \log p \left( \mathbf{z} \right) \\
    & \quad +\frac{\rho}{2} \left\| \mathbf{x} - \mathbf{z} + \mathbf{u} \right\|_2^2 - \frac{\rho}{2} \left\| \mathbf{u} \right\|_2^2, 
\end{aligned}
\end{equation}
where $\mathbf{u}$ is the dual variable and $\rho$ is a hyperparameter that defines the strength of the soft constraints. ADMM then applies an alternating gradient descent approach to minimizing the Augmented Lagrangian, resulting in a set of updates on $\mathbf{x,z,u}$ that are applied in an iterative fashion:
%
\begin{align}
\mathbf{x} & \leftarrow \operatorname*{argmin}_{\mathbf{x}} \frac{1}{2 \sigma^2} \left\| \mathbf{y} - \mathcal{A} \left( \mathbf{x} \right) \right\|_2^2 + \frac{\rho}{2} \| \mathbf{x} - \mathbf{z} + \mathbf{u} \|_2^2 \label{eq:admm_updates:1} \\
    \mathbf{z} &\leftarrow \operatorname*{argmin}_{\mathbf{z}} - \log p \left( \mathbf{z} \right) + \frac{\rho}{2} \| \mathbf{x} - \mathbf{z} + \mathbf{u} \|_2^2 \\
    &= \mathcal{D} \left(\mathbf{x} + \mathbf{u}, \tilde{\sigma}^2 = \frac{1}{\rho} \right) \label{eq:admm_updates:2} \\
    \mathbf{u} &\leftarrow \mathbf{u} + \mathbf{x} - \mathbf{z}. \label{eq:admm_updates:3}
\end{align}

Here, the $\mathbf{x}$-update is an unconstrained least-squares problem that does not depend on the prior and which often has a closed-form, or at least an efficient, solution. An important insight of plug-and-play image restoration methods~\cite{pnp,chan2016plug} is the fact that the $\mathbf{z}$-update (Eq.~\ref{eq:admm_updates:2}) is a denoising problem on the variable $\mathbf{x}+\mathbf{u}$, which can be solved using any Gaussian denoiser $\mathcal{D} \left(\cdot, \tilde{\sigma}^2 \right)$ assuming the noise level is $\tilde{\sigma}$.

Dual ascent, for example implemented by ADMM,  offers several benefits in traditional optimization, including the ability to leverage convexity in the dual problem, leading to simpler, more efficient, and more robust optimization methods~\cite{admm}.

\begin{figure*}[htbp]
    \centering
    \includegraphics[width=\textwidth]{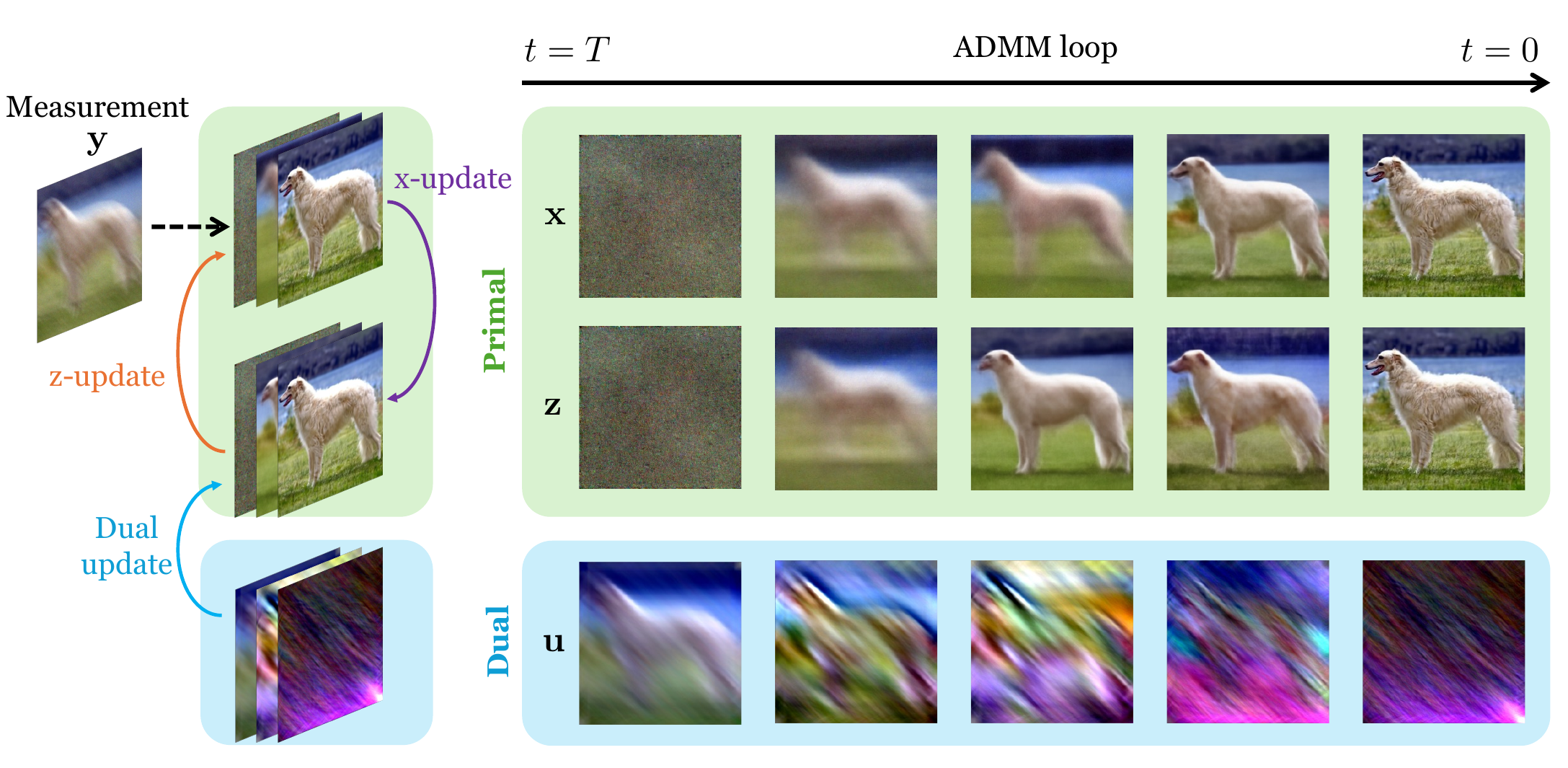}
    \caption{\textbf{Overview of \method.} This example illustrates motion deblurring. \Method\ alternates between three updates ($\mathbf{x}$-, $\mathbf{z}$-, and dual updates) within each ADMM iteration. As iterations proceed, the primal variables $\mathbf{x}$ and $\mathbf{z}$ progressively align, while the dual variable $\mathbf{u}$ diminishes toward zero, indicating convergence to a fixed point $(\mathbf{x}^\star, \mathbf{z}^\star, \mathbf{u}^\star)$. This evolution visually demonstrates the fixed-point convergence behavior analyzed in Appendix~\ref{appendix:convergence}. Here, the ADMM iteration index and diffusion timestep are set equal by design.}
    
    \label{fig:algo_diagram}
\end{figure*}

\subsection{Diffusion Models and Posterior Sampling}
\label{sec:diffusion}

The key insight of diffusion models lies in the fact that one can sample from a target distribution $p_0(\mathbf{x})$ by first sampling $\mathbf{x}_T$ from another distribution $p_T$ that is easy to sample from, e.g., a Gaussian, and iteratively applying a \emph{reverse} diffusion step of the form
\begin{equation}
    \mathbf{x}_{t-1} = \frac{1}{\sqrt{{\alpha}_t}} \left( \mathbf{x}_t + \left( 1 - {\alpha}_t \right) \mathbf{s}_\theta(\mathbf{x}_t, t) \right) + \sqrt{1 - {\alpha}_t} \,\epsilon
    \label{eq:reverse_diffusion}
\end{equation}
for $t=T, \ldots, 1$ to generate an image $\mathbf{x}_0$, where $\epsilon \sim \mathcal{N} \left( \mathbf{0}, \mathbf{I} \right)$. The reverse diffusion process~\cite{anderson1982reverse, haussmann1986time} approximates the inverse trajectories of a corresponding \emph{forward} diffusion $\mathbf{x}_t = \sqrt{\bar{\alpha}}_t \, \mathbf{x}_0 + \sqrt{1 - \bar{\alpha}_t} \, \epsilon$, $\epsilon \sim \mathcal{N} \left( \mathbf{0}, \mathbf{I} \right)$. Here, we adopt the variance-preserving form of forward and reverse diffusion~\cite{ho2020denoising}. The factors $\alpha_t$ and $\bar{\alpha}_t=\prod_{s=0}^t\alpha_s$ are derived from the noise schedule of the diffusion model~\cite{ho2020denoising}. Importantly, the score network $\mathbf{s}_\theta(\mathbf{x}_t, t)$, defined by parameters $\theta$, is a neural network that approximates the \emph{score function} $\nabla_{\mathbf{x}}\log p_t(\mathbf{x})$~\cite{song2021maximum}. This network is learned from training data in the diffusion model pretraining stage. To sample more efficiently, Denoising Diffusion Implicit Models (DDIM) \cite{ddim} provides an alternative non-Markovian reverse parameterization of the diffusion process, replacing the Markovian formulation in Eq.~\ref{eq:reverse_diffusion}:
\begin{equation}
\begin{aligned}
    \mathbf{x}_{t-1} &= \sqrt{\bar{\alpha}_{t-1}} \left( \frac{\mathbf{x}_t + (1-\bar{\alpha}_t)\mathbf{s}_\theta(\mathbf{x}_t,t)}{\sqrt{\bar{\alpha}_t}} \right) \\
    &\quad- \sqrt{1-\bar{\alpha}_{t-1}-\sigma_t^2} \, \left(\sqrt{1-\bar{\alpha}_t}\cdot\mathbf{s}_\theta(\mathbf{x}_t,t)\right) \, + \sigma_t\epsilon,
    \label{eq:reverse_diffusion_ddim}
\end{aligned}
\end{equation}
where $\epsilon \sim \mathcal{N} \left( \mathbf{0}, \mathbf{I} \right)$ and $\sigma_t$ is a schedule chosen at inference.

In a posterior sampling problem, we aim at sampling from the posterior $p(\mathbf{x}|\mathbf{y})$. For this purpose, many recent methods follow the approach described above, using the \textit{posterior score} $\nabla_\mathbf{x}\log p_t(\mathbf{x}|\mathbf{y}) = \nabla_\mathbf{x}\log p(\mathbf{y}|\mathbf{x}_t=\mathbf{x}) + \nabla_\mathbf{x}\log p_t(\mathbf{x})$ instead of $\nabla_\mathbf{x}\log p_t(\mathbf{x})$. 
The second term is equivalent to the unconditional score of the pretrained diffusion model, but the challenge lies in the first, i.e., the conditional score term. The conditional probability $p(\mathbf{y}|\mathbf{x}_t)$ can be written as a conditional expectation $\mathbb{E}_{\mathbf{x}_0\sim p(\mathbf{x}_0|\mathbf{x}_t)}[ p(\mathbf{y}|\mathbf{x}_0)]$, but approximating this expectation with Monte Carlo samples is computationally intractable (see \citet{dps}, for example). For this reason, existing diffusion posterior sampling methods approximate this conditional distribution with a Dirac delta distribution concentrated on $\mathbf{x}_t$~\cite{score-ald} or on $\mathbb{E}_{\mathbf{x}_0\sim p(\mathbf{x}_0|\mathbf{x}_t)}[\mathbf{x}_0]$~\cite{dps}. A number of other diffusion posterior sample methods have been proposed~\cite{survey}, each providing a different approximation for the conditional expectation. Most recently, \citet{daps} introduced DAPS (Decoupled Annealing Posterior Sampling), a two-step iterative approach that mitigates the accumulation of errors along the sampling trajectory through a Markov chain Monte Carlo--based equilibration step at $t=0$. DAPS demonstrated state-of-the-art results on both linear and nonlinear inverse problems, but remains limited by the necessity to approximate the conditional probability $p(\mathbf{x}_0|\mathbf{x}_t)$ and by the finiteness of the number of MCMC steps.

\section{Method}

We derive our approach to solving MAP problems with pretrained diffusion model priors in the following. 


\subsection{Diffusion Plug-and-Play ADMM for Image Restoration}


We dub the naive approach for using ADMM with a pretrained diffusion model \emph{Diff-PnP-ADMM}. For this purpose, we apply the ADMM framework as discussed in Sec.~\ref{sec:map} and use the pretrained diffusion model as a one-step denoiser $\mathcal{D}$ in the $\mathbf{z}$-update (i.e., Eq.~\ref{eq:admm_updates:2}). This is done by applying Tweedie's formula~\cite{efron2011tweedie} and replacing the $\mathbf{z}$-update with
%
\begin{equation}
\mathbf{z}\leftarrow\frac{1}{\sqrt{\bar{\alpha}_t}}\left(\mathbf{x}+\mathbf{u}+(1-\bar{\alpha}_t)\mathbf{s}_\theta(\mathbf{x}+\mathbf{u},t)\right)
\label{diff-pnp-admm}
\end{equation}
%
%
where $t$ decreases at each step of the ADMM loop, which implicitly constrains the relationship between the diffusion schedule $\alpha_t$ and the soft constraint parameter $\rho$.


\subsection{Dual Ascent Diffusion (\Method)}



At its core, the ADMM method iterates over 3 steps. The $\mathbf{x}$-update corresponds to a data matching step (Eq.~\ref{eq:admm_updates:1}), the $\mathbf{z}$-update can be seen as a denoising step (Eq.~\ref{eq:admm_updates:2}) and the dual update stems from the introduction of the dual variable $\mathbf{u}$ for the constraint $\mathbf{x}=\mathbf{z}$ (Eq.~\ref{eq:admm_updates:3}). While the $\mathbf{x}$ and dual updates are straightforwardly derived from the original ADMM framework, our methodological contribution primarily consists of showing that a pretrained diffusion model can be used more efficiently in the $\mathbf{z}$-update than the naive approach (Eq.~\ref{diff-pnp-admm}).

\textbf{$\mathbf{x}$-update.} In order for the method to be directly applicable to any differentiable forward model $\mathcal{A}$, whether linear or nonlinear, we replace the minimization problem of Eq.~\ref{eq:admm_updates:1} with a single gradient step 
\begin{equation}
    \mathbf{x}\leftarrow \mathbf{v} -\gamma\nabla_{\mathbf{v}}\Vert\mathbf{y}-\mathcal{A}(\mathbf{v})\Vert_2^2, \quad \mathbf{v} =\mathbf{z}-\mathbf{u},
    \label{eq:x-update}
\end{equation}
where $\gamma$ is a step size that can be adjusted at each iteration. This approach is also known as linearized ADMM~\cite{parikh2014proximal}.

\textbf{$\mathbf{u}$-update.} The update of the dual variable is readily available in Eq.~\ref{eq:admm_updates:3}.

\textbf{$\mathbf{z}$-update.} Following Eq.~\ref{eq:admm_updates:2}, the $\mathbf{z}$-update consists in denoising $\mathbf{x}+\mathbf{u}$ for a certain noise level $\tilde{\sigma}^2$. However, it is crucial to note that since the score model $\mathbf{s}_\theta(\mathbf{x},t)$ is only trained on points sampled from $p_t$, it is a poor approximation of the true score whenever $\mathbf{x}$ is unlikely under $p_t$ (i.e., when $p_t(\mathbf{x})\ll 1$). In other words, the score model at time $t$ is only accurate on points belonging to the diffusion manifold at time $t$~\cite{manifold-constraint}. Because $\mathbf{x}+\mathbf{u}$ does not in general belong to this manifold, we propose to replace the $\mathbf{z}$-update with
\begin{equation}
    \mathbf{z}\leftarrow \frac{1}{\sqrt{\bar{\alpha}_t}} \left(\mathbf{x}_t+(1-\bar{\alpha}_t)\mathbf{s}_\theta(\mathbf{x}_t,t) \right), \label{eq:denoising}
\end{equation}
where $\mathbf{x}_t$ is defined recursively following $\mathbf{x}_T\sim\mathcal{N}(\mathbf{0},\mathbf{I})$ and
\begin{equation}
    \mathbf{x}_{t-1} \leftarrow \underbrace{\sqrt{\bar{\alpha}_{t-1}}\cdot \mathbf{x} + \sqrt{1-\bar{\alpha}_{t-1}-\sigma_t^2} \cdot \hat{\epsilon} + \sigma_t \epsilon}_{\text{DDIM update}}+\underbrace{\sqrt{\bar{\alpha}_{t-1}}\cdot\mathbf{u}}_{\text{Re-scaled }\mathbf{u}}.
    \label{eq:forward}
\end{equation}
In this equation, $\hat{\epsilon}= \left( \mathbf{x}_t - \sqrt{\bar{\alpha}_t}\cdot\mathbf{x}  \right)/\sqrt{1-\bar{\alpha}_t}$ and $\sigma_t$ is a hyperparameter. The first part of the right-hand side corresponds to the DDIM update where the ``predicted $\mathbf{x}_0$'' is $\mathbf{x}$. The second part adds the dual variable $\mathbf{u}$, re-scaled to match the signal level of $\mathbf{x}_{t-1}$.

Combined iteratively, these three steps define our method (Algorithm~\ref{alg:admm_ddim_algo}). We show that the DDiff updates converge to a fixed point $(\mathbf{x}^\star, \mathbf{z}^\star, \mathbf{u}^\star)$: under bounded denoiser and gradient assumptions, all three iterate sequences are Cauchy and converge to a limit (Theorem~\ref{thm:fp-ddiff}); notably, this result does not require a convex prior. Full assumptions, proofs, and technical details are in Appendix~\ref{appendix:convergence}.

\subsection{DDiff with Latent Diffusion Model}

Given a pretrained encoder $\mathcal{E}$ and decoder $\mathcal{D}$, latent diffusion models (LDMs)~\cite{ldm} are trained to model the distribution $p(\mathbf{z}_0)$ of latent variables $\mathbf{z}_0 = \mathcal{E}(\mathbf{x}_0)$, enabling reconstruction via $\mathbf{x}_0 = \mathcal{D}(\mathbf{z}_0)$. Leveraging latent diffusion models typically offers reduced memory and compute demands compared to pixel-space diffusion. Extending \Method\ to this latent domain yields LatentDDiff, which performs alternating data-fidelity and denoising updates on compact latent representations while preserving the dual-ascent structure of \Method. The full algorithmic formulation, quantitative results, and discussion of enforcing data consistency either in latent or pixel space are provided in Appendix~\ref{sec:latent_ddiff_supp}.


\subsection{Comparison to Other Diffusion-based Variable Splitting Methods}
Prior methods, such as DiffPIR \cite{diffpir}, DCDP \cite{dcdp}, and PnP-DM \cite{wu-principled}, are MAP-based optimization methods, specifically built upon the half-quadratic splitting (HQS) method~\cite{hqs}, which lacks dual variables in its formulation. Our work naturally extends existing frameworks, transforming them from HQS-style approaches to those using dual variables, including ADMM. By incorporating Lagrange multipliers that accumulate constraint violations across the iterations, \Method offers improved empirical performance on challenging inverse problems where measurement consistency is crucial. In particular, we note that removing the dual update from \method would exactly emulate DiffPIR~\cite{diffpir} (with the right choice of $\sigma_t$ and $\gamma_t$, see Appendix~\ref{appendix:relation}).


\begin{algorithm}
\caption{\Method}
\label{alg:admm_ddim_algo}
\begin{algorithmic}[1]  
\REQUIRE $T$, $\mathcal{A}(\cdot)$, $\{ \sigma_t \}_{t=1}^{T}$, $\{ \bar{\alpha}_t \}_{t=1}^{T}$, $\mathbf{s}_\theta$,  $\mathbf{y}$, $\{\gamma_t\}_{t=1}^T$, $t_0$

\STATE Initialize $\mathbf{x}_T \sim \mathcal{N}(\mathbf{0}, \mathbf{I}), \mathbf{u} = \mathbf{0} $.
\FOR{$t = T-1$ $\textbf{to}$ $0$}
    \STATE $\mathbf{z} \leftarrow \frac{1}{\sqrt{\bar{\alpha}_t}} \left(\mathbf{x}_t + (1-\bar{\alpha}_t) \mathbf{s}_\theta(\mathbf{x}_t, t)\right)$ 
    \STATE \hfill {\color{gray} $\triangleright$ \textit{Denoising step (Eq.~\ref{eq:denoising})}}
    \STATE $\mathbf{x} \leftarrow  \mathbf{z} - \mathbf{u} - \gamma_t \nabla_{\mathbf{v}=\mathbf{z} - \mathbf{u}} \|\mathbf{y} - \mathcal{A}( \mathbf{v} )\|^2$ \STATE \hfill {\color{gray}$\triangleright$ \textit{Measurement step (Eq.~\ref{eq:x-update})}}
    \STATE $\hat{\epsilon}\leftarrow \frac{1}{\sqrt{1-\bar{\alpha}_t}} \left( \mathbf{x}_t - \sqrt{\bar{\alpha}_t}\cdot\mathbf{x}  \right)$
    \STATE $\epsilon \sim \mathcal{N}(\mathbf{0}, \mathbf{I}) \:\textbf{if} \:t>t_0\:\textbf{else} \: \epsilon=0$
    \STATE $\mathbf{x}_{t-1} \leftarrow\sqrt{\bar{\alpha}_{t-1}} \cdot\mathbf{x} + \sqrt{1-\bar{\alpha}_{t-1}-\sigma_t^2} \cdot \hat{\epsilon} + \sigma_t \epsilon+\sqrt{\bar{\alpha}_{t-1}}\cdot\mathbf{u}$ \hfill {\color{gray} $\triangleright$ \textit{Reverse diffusion (Eq.~\ref{eq:forward})}}
    \STATE $\mathbf{u} \leftarrow \mathbf{u} + \mathbf{x} - \mathbf{z}$\hfill {\color{gray} $\triangleright$ \textit{Dual update (Eq.~\ref{eq:admm_updates:3})}}
\ENDFOR

\RETURN $\mathbf{x}_0$
\end{algorithmic}
\end{algorithm}
\section{Experiments}
\label{sec:experimental_setup}
\subsection{Experimental Setup}

\paragraph{Datasets and metrics.} We evaluate our method on two image datasets, FFHQ $256 \times 256$ \cite{ffhq} and ImageNet $256 \times 256$ \cite{imagenet}. For pixel-space diffusion models, we utilize pretrained models from \cite{dps} on the FFHQ dataset and from \cite{imagenet_model} on the ImageNet dataset. For latent diffusion models, we used unconditional LDM-VQ4 (autoencoder with a downsampling factor of 4) pretrained models from \cite{resample} for FFHQ and \cite{ldm} for ImageNet. We randomly selected 100 images from the validation set for both datasets, and the images were normalized to $[-1,1]$. Our main evaluation metrics include peak signal-to-noise ratio (PSNR), structural similarity index measure (SSIM), learned perceptual image patch similarity (LPIPS) \cite{lpips}, and residual error, defined as $\Vert\mathbf{y}-\mathcal{A}(\mathbf{x})\Vert_2^2-\sigma^2$ (see Appendix~\ref{appendix:residual} for details), which quantifies the degree of data consistency.


\paragraph{Inverse problems.} Our method is evaluated on multiple inverse problems. \textbf{Linear tasks} include super-resolution ($4\times$ downsampling), Gaussian deblurring, motion deblurring, inpainting with a $128\times128$ box, and inpainting with a random mask that removes $70\%$ of the pixels. \textbf{Nonlinear tasks} consist of phase retrieval (oversampling ratio of $2.0$), for which we report the best result out of five runs due to the intrinsic instability of the task, nonlinear deblurring, and high dynamic range ($2\times$ dynamic range). All measurements include additive Gaussian noise ($\sigma=0.05$). 

\begin{figure*}[htbp]
  \centering
  \includegraphics[height=0.95\textheight, keepaspectratio]{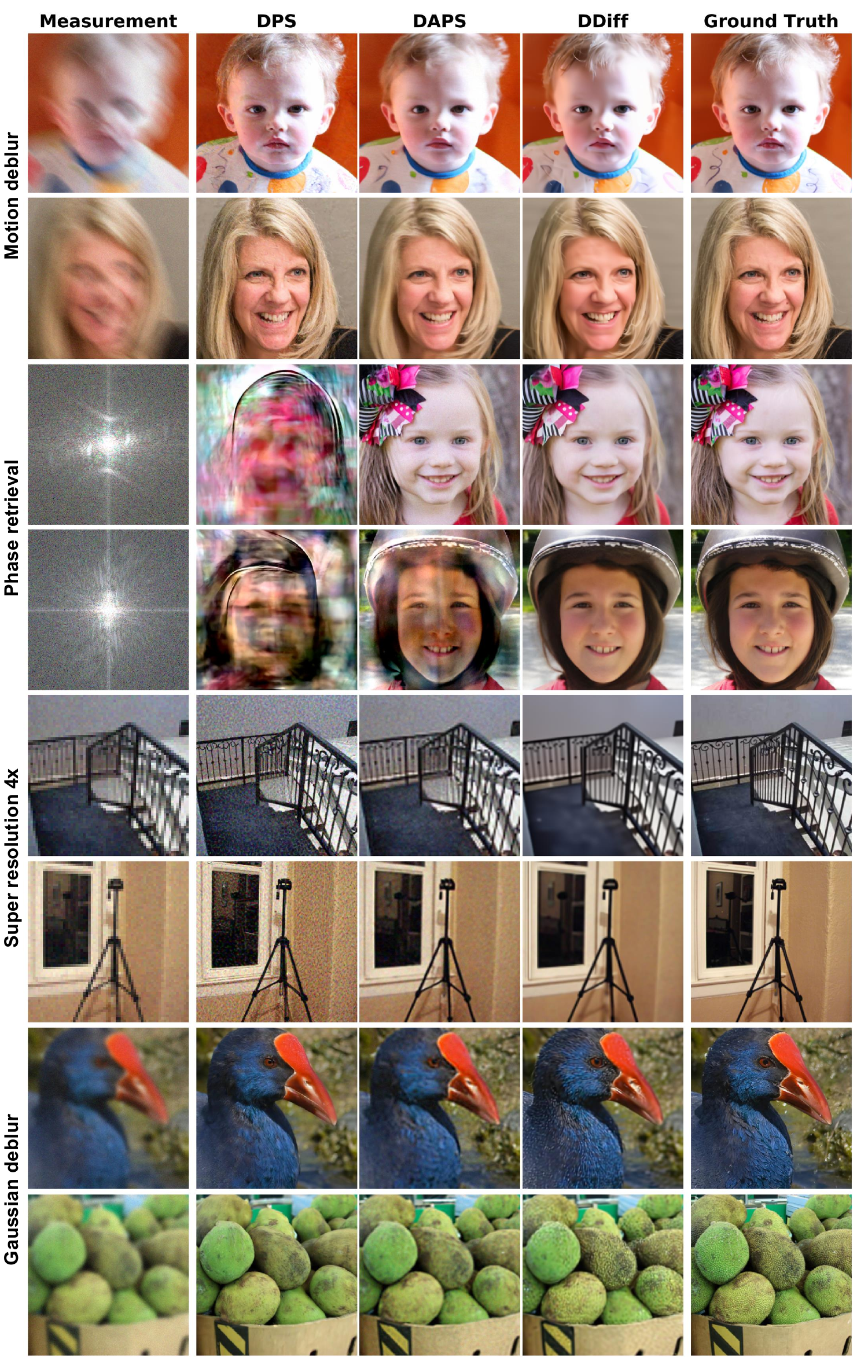}
  \caption{\textbf{Qualitative results.} \Method demonstrates sharper and cleaner results compared to DPS \cite{dps} and DAPS \cite{daps}. All tasks are run with a noise of standard deviation $\sigma=0.05$.}
  \label{fig:main_results}
\end{figure*}

\begin{table*}[htbp]
\caption{\textbf{Quantitative evaluation.} Comparing different methods for 5 linear and 3 nonlinear tasks on FFHQ and ImageNet datasets. This evaluation uses 100 validation images and reports the average metric value. The best and second-best results are distinguished by \textbf{bold} and \underline{underlined} marks, respectively. All tasks are run with a noise of standard deviation $\sigma=0.05$. See Appendix~\ref{appendix:stat_sig} for 95\% confidence intervals.}
  \centering
  \begin{adjustbox}{center, max width=\textwidth, max totalheight=\textheight, keepaspectratio}
  \scriptsize
  \begin{tabular}{l|l|cccc|cccc}
    \toprule
    & & \multicolumn{4}{c|}{\textbf{FFHQ}} & \multicolumn{4}{c}{\textbf{ImageNet}} \\
    \textbf{Task} & \textbf{Method} & PSNR ($\uparrow$) & SSIM ($\uparrow$) & LPIPS ($\downarrow$) & Residual ($\downarrow$) & PSNR ($\uparrow$) & SSIM ($\uparrow$) & LPIPS ($\downarrow$) & Residual ($\downarrow$) \\
    \midrule
    
    \multirow{4}{*}{\begin{tabular}[c]{@{}l@{}}Super\\Resolution 4×\end{tabular}} 
    & \Method (ours) & \textbf{30.07} & \textbf{0.824} & \underline{0.211} & \textbf{0.0028} & \textbf{25.81} & \textbf{0.656} & \underline{0.396} & \textbf{0.0038} \\
    & DAPS & \underline{29.34} & \underline{0.783} & \textbf{0.190} & \underline{0.0029} & \underline{25.44} & \underline{0.636} & \textbf{0.295} & 0.0047 \\
    & DMPlug & 28.55 & 0.742 & 0.220 & 0.0038 & 24.22 & 0.649 & 0.432 & \underline{0.0039} \\
    & DPS & 24.42 & 0.486 & 0.346 & 0.0050 & 21.10 & 0.351 & 0.408 & 0.0052 \\
    & DiffPIR & 23.71 & 0.440 & 0.423 & 0.0087 & 20.75 & 0.312 & 0.517 & 0.0093 \\
    & DCDP & 26.65 & 0.641 & 0.410 & 0.0079 & 23.51 & 0.542 & 0.460 & 0.0060 \\
    & DDRM & 26.32 & 0.763 & 0.286 & 0.0075 & 22.26 & 0.513 & 0.473 & 0.0072 \\
    \midrule
    
    \multirow{4}{*}{\begin{tabular}[c]{@{}l@{}}Inpainting\\(Box)\end{tabular}} 
    & \Method (ours) & \textbf{24.88} & \textbf{0.831} & \underline{0.110} & \underline{0.0077} & \underline{21.15} & \textbf{0.743} & \underline{0.240} & \textbf{0.0119} \\
    & DAPS & \underline{24.12} & 0.742 & 0.174 & 0.0099 & \textbf{21.22} & 0.714 & \textbf{0.230} & 0.0150 \\
    & DPS & 23.68 & \underline{0.810} & \textbf{0.079} & \textbf{0.0033} & 19.63 & 0.725 & 0.254 & 0.0412 \\
    & DiffPIR & 19.02 & 0.527 & 0.252 & 0.0106 & 16.02 & 0.520 & 0.329 & 0.0145 \\
    & DCDP & 23.67 & 0.729 & 0.232 & 0.0101 & 20.45 & \underline{0.732} & 0.248 & \underline{0.0132} \\
    & DDRM & 22.15 & 0.701 & 0.209 & 0.0099 & 18.52 & 0.713 & 0.254 & 0.0194 \\
    & RED-diff & 14.57 & 0.578 & 0.586 & 0.0451 & 13.98 & 0.628 & 0.367 & 0.0437 \\
    \midrule
    
    \multirow{4}{*}{\begin{tabular}[c]{@{}l@{}}Inpainting\\(Random)\end{tabular}} 
    & \Method (ours) & \textbf{33.08} & \textbf{0.877} & \textbf{0.050} & \textbf{0.0205} & \textbf{28.39} & \textbf{0.758} & \textbf{0.136} & \textbf{0.0241} \\
    & DAPS & 30.76 & 0.801 & 0.156 & 0.0293 & \underline{27.32} & 0.725 & \underline{0.189} & 0.0788 \\
    & DMPlug & \underline{31.65} & \underline{0.852} & 0.137 & 0.0290 & 26.09 & \underline{0.740} & 0.245 & \underline{0.0316} \\
    & DPS & 30.79 & 0.807 & \underline{0.083} & \underline{0.0217} & 27.31 & 0.737 & 0.235 & 0.0980 \\
    & DiffPIR & 18.53 & 0.362 & 0.622 & 0.0264 & 15.82 & 0.191 & 0.842 & 0.1080 \\
    & DCDP & 25.67 & 0.757 & 0.224 & 0.0398 & 20.40 & 0.723 & 0.253 & 0.1001 \\
    \midrule

    \multirow{4}{*}{\begin{tabular}[c]{@{}l@{}}Gaussian\\Deblurring\end{tabular}} 
    & \Method (ours) & \underline{28.87} & \textbf{0.800} & \textbf{0.119} & \textbf{0.0026} & \underline{22.29} & 0.471 & 0.415 & \textbf{0.0046} \\
    & DAPS & \textbf{29.63} & \underline{0.789} & 0.177 & \underline{0.0027} & \textbf{25.90} & \textbf{0.658} & \textbf{0.269} & 0.0084 \\
    & DMPlug & 22.98 & 0.537 & 0.288 & 0.0036 & 14.82 & 0.188 & 0.680 & 0.0147 \\
    & DPS & 27.77 & 0.704 & \underline{0.140} & 0.0029 & 21.07 & \underline{0.528} & \underline{0.392} & \underline{0.0073} \\
    & DiffPIR & 26.16 & 0.624 & 0.297 & 0.0031 & 21.64 & 0.393 & 0.497 & 0.0093 \\
    & DCDP & 16.75 & 0.173 & 0.701 & 0.0141 & 16.06 & 0.183 & 0.674 & 0.0195 \\
    & DDRM & 24.87 & 0.725 & 0.246 & 0.0052 & 21.14 & 0.457 & 0.464 & 0.0113 \\
    & RED-diff & 12.18 & 0.149 & 1.232 & 0.0348 & 12.22 & 0.128 & 0.613 & 0.0256 \\
    \midrule

    \multirow{4}{*}{\begin{tabular}[c]{@{}l@{}}Motion\\Deblurring\end{tabular}} 
    & \Method (ours) & \underline{28.24} & \underline{0.785} & \textbf{0.129} & \textbf{0.0058} & \underline{24.16} & 0.585 & \underline{0.242} & \textbf{0.0079} \\
    & DAPS & \textbf{29.17} & \textbf{0.797} & 0.186 & \underline{0.0059} & \textbf{26.61} & \textbf{0.710} & \textbf{0.241} & 0.0085 \\
    & DMPlug & 21.95 & 0.512 & 0.304 & 0.0076 & 14.81 & 0.170 & 0.696 & 0.0199 \\
    & DPS & 27.93 & 0.714 & \underline{0.130} & 0.0061 & 23.36 & \underline{0.611} & 0.321 & \underline{0.0082} \\
    & DiffPIR & 22.01 & 0.327 & 0.499 & 0.0074 & 18.93 & 0.248 & 0.586 & 0.0084 \\
    & DCDP & 9.536 & 0.039 & 0.855 & 0.0547 & 9.491 & 0.066 & 0.771 & 0.0511 \\
    \midrule

    \multirow{4}{*}{\begin{tabular}[c]{@{}l@{}}Phase \\Retrieval\end{tabular}} 
    & \Method (ours) & \textbf{29.94} & \textbf{0.816} & \textbf{0.120} & \textbf{0.0040} & \underline{18.54} & \textbf{0.494} & \textbf{0.262} & \textbf{0.0059} \\
    & DAPS & \underline{29.60} & \underline{0.768} & \underline{0.182} & \underline{0.0049} & \textbf{20.23} & \underline{0.449} & \underline{0.397} & \underline{0.0085} \\
    & DPS & 22.24 & 0.540 & 0.307 & 0.0514 & 16.03 & 0.396 & 0.444 & 0.1040 \\
    & DiffPIR & 10.04 & 0.036 & 0.783 & 0.1811 & 9.61 & 0.021 & 0.794 & 0.2410 \\
    & DCDP & 15.20 & 0.420 & 0.616 & 0.1060 & 11.63 & 0.201 & 0.700 & 0.1220 \\
    & RED-diff & 14.88 & 0.386 & 0.656 & 0.1721 & 13.89 & 0.266 & 0.639 & 0.1121 \\
    \midrule

    \multirow{4}{*}{\begin{tabular}[c]{@{}l@{}}Nonlinear\\Deblurring\end{tabular}} 
    & \Method (ours) & \textbf{31.48} & \textbf{0.873} & \textbf{0.120} & \textbf{0.0027} & \textbf{29.68} & \textbf{0.805} & \textbf{0.207} & \textbf{0.0035} \\
    & DAPS & 28.45 & 0.764 & 0.188 & 0.0042 & 27.28 & \underline{0.718} & \underline{0.213} & \underline{0.0048} \\
    & DMPlug & 27.17 & 0.791 & 0.187 & 0.0051 & 22.99 & 0.603 & 0.366 & 0.0104 \\
    & DPS & 25.39 & 0.643 & 0.258 & 0.0095 & \underline{28.42} & 0.691 & 0.271 & 0.0053 \\
    & DiffPIR & 19.79 & 0.331 & 0.583 & 0.0273 & 22.13 & 0.459 & 0.435 & 0.0167 \\
    & DCDP & \underline{28.87} & \underline{0.852} & \underline{0.177} & \underline{0.0038} & 25.22 & 0.700 & 0.299 & 0.0105 \\
    & RED-diff & 29.89 & 0.783 & 0.185 & 0.0040 & 28.07 & 0.624 & 0.306 & 0.0051 \\
    \midrule

    \multirow{4}{*}{\begin{tabular}[c]{@{}l@{}}High\\Dynamic\\Range\end{tabular}} 
    & \Method (ours) & \underline{26.05} & \textbf{0.873} & \textbf{0.129} & \textbf{0.0459} & \textbf{26.50} & \underline{0.800} & \textbf{0.108} & \textbf{0.0541} \\
    & DAPS & \textbf{27.39} & \underline{0.846} & \underline{0.163} & \underline{0.0505} & \underline{26.10} & \textbf{0.825} & \underline{0.171} & \underline{0.0717} \\
    & DPS & 25.79 & 0.793 & 0.165 & 0.0734 & 22.72 & 0.721 & 0.273 & 0.1951 \\
    & DiffPIR & 17.69 & 0.645 & 0.296 & 0.1292 & 18.23 & 0.637 & 0.289 & 0.2105 \\
    & RED-diff & 21.28 & 0.431 & 0.359 & 0.0921 & 21.02 & 0.567 & 0.479 & 0.1322 \\
    \bottomrule
  \end{tabular}
  \end{adjustbox}
  \label{tab:comparison}
\end{table*}

\paragraph{Baselines.} We compare against the state of the art: DAPS \cite{daps}, DMPlug \cite{dmplug}, DCDP \cite{dcdp}, RED-diff \cite{red-diff}, DDRM \cite{ddrm}, DPS \cite{dps}, and DiffPIR \cite{diffpir}. Notably, DiffPIR and DDRM were not proven to handle nonlinear tasks. We closely evaluate our method against DAPS as it achieves the best results among the baselines. For latent diffusion adaptation of our method, we compare against LatentDAPS \cite{daps}, PSLD \cite{psld}, and ReSample \cite{resample}. Note that PSLD cannot handle nonlinear inverse tasks.

\begin{figure*}[htbp]
    \centering
    \begin{subfigure}[b]{0.49\textwidth}
        \centering
        \includegraphics[width=\linewidth]{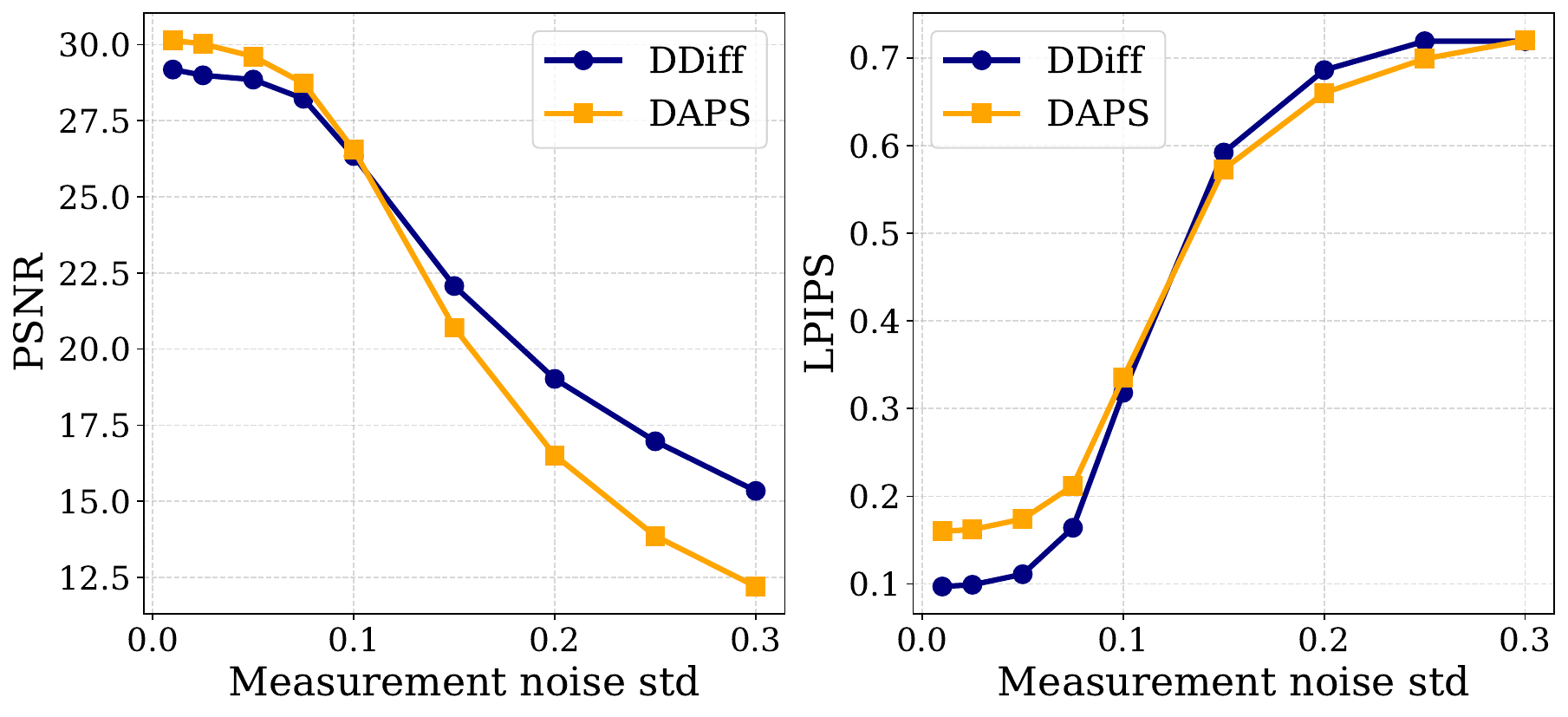}
        \caption{Gaussian deblurring}
        \label{fig:deblur_high_noise}
    \end{subfigure}
    \hspace{0.1mm}
    \begin{subfigure}[b]{0.49\textwidth}
        \centering
        \includegraphics[width=\linewidth]{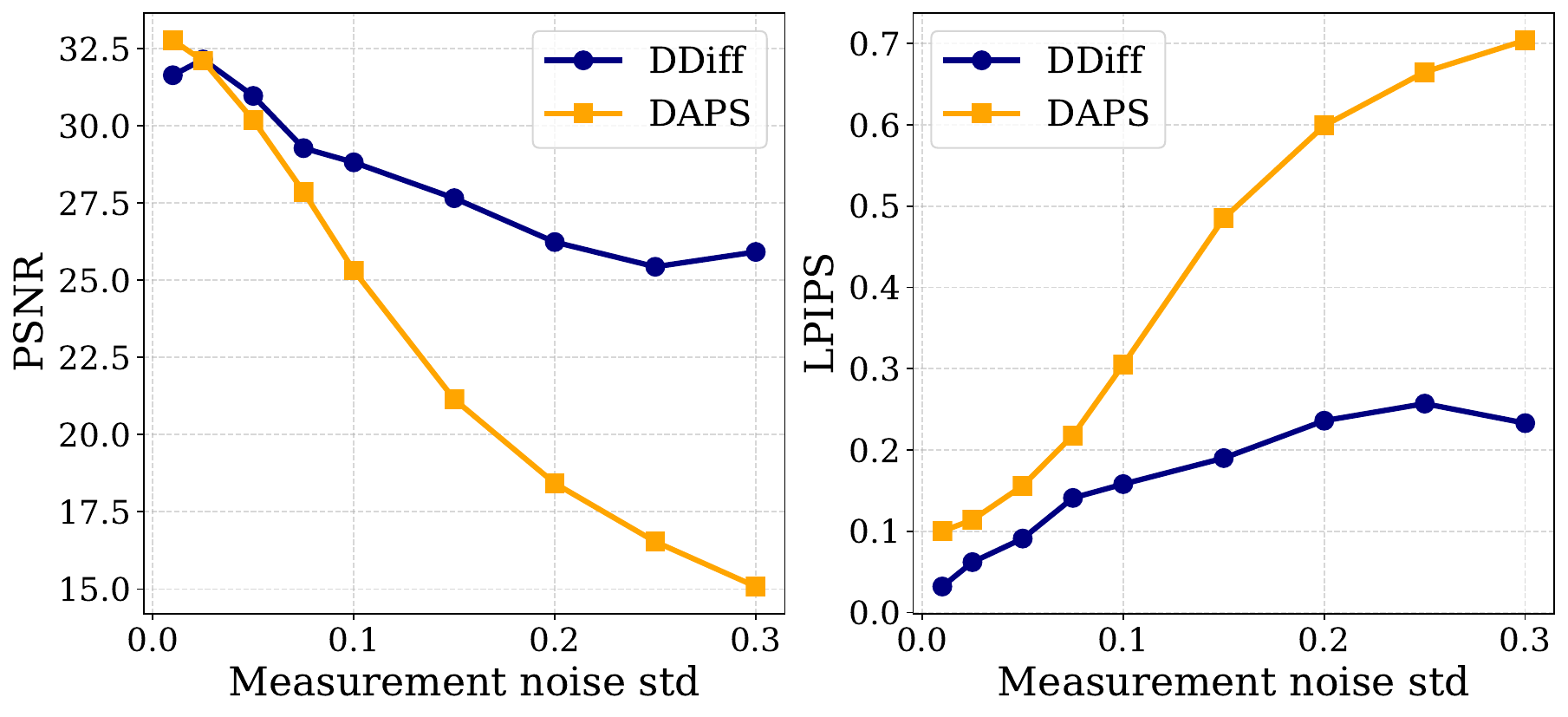}
        \caption{Phase retrieval}
        \label{fig:pr_high_noise}
    \end{subfigure}
    \caption{\textbf{Effect of measurement noise level.} \Method demonstrates greater robustness as noise increases. This evaluation uses 10 FFHQ validation images.}
    \label{fig:high_meas}
\end{figure*}

\begin{figure*}[htbp]
  \centering
  \includegraphics[width=\textwidth]{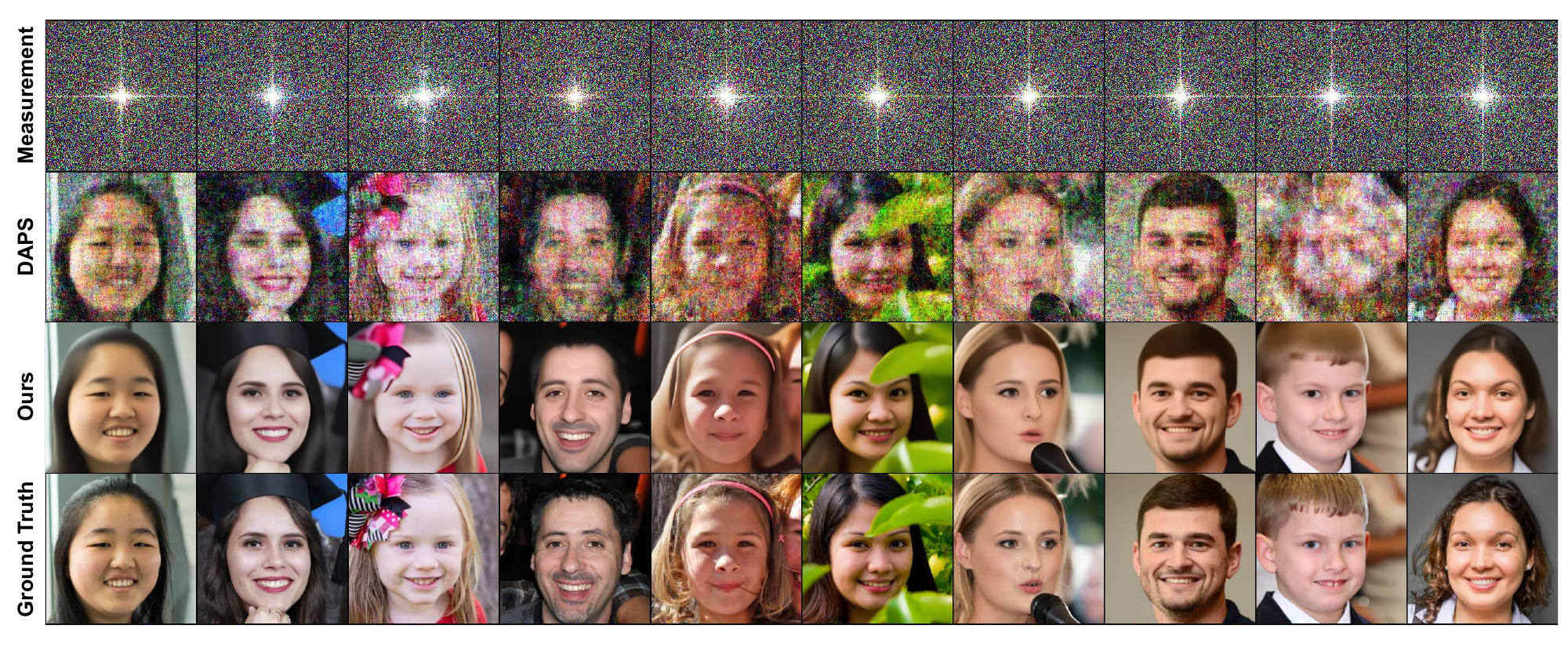}
  \caption{\textbf{Qualitative results at high measurement noise level.} We compare our method (\Method) with DAPS~\cite{daps} on the phase retrieval task at $\sigma = 0.3$, using 10 randomly selected validation images. Despite the severe measurement noise, \Method\ successfully recovers coherent global structures and facial semantics, whereas DAPS reconstructions exhibit strong corruption and noise artifacts.}
  \label{fig:high_meas_grid}
\end{figure*}

\subsection{Main Results}
Quantitative evaluation for the linear and nonlinear tasks on FFHQ and ImageNet datasets are shown in Table~\ref{tab:comparison}. Our method outperforms the baselines on the vast majority of the tasks, especially in terms of perceptual similarity and residual error.
This is further demonstrated in Fig.~\ref{fig:main_results}, where we show a qualitative comparison between the baselines and our method. Overall, \method reconstructs finer details with fewer visual artifacts. Additional qualitative examples and hyperparameter details are provided in the supplements.

Moreover, \method exhibits significantly increased robustness to higher measurement noise. As shown in Fig.~\ref{fig:high_meas}, although PSNR may be slightly lower than DAPS in the very low measurement noise region (approximately $\sigma<0.05$), PSNR degrades less rapidly as $\sigma$ increases.
LPIPS is generally lower for \Method than that of DAPS in all levels of noise. 
LPIPS follows a similar trend, especially on the phase retrieval task, where \method divides LPIPS by 3 in the high-noise region ($\sigma = 0.3$).
To visualize the reconstruction quality of \Method compared to DAPS at the extreme case of measurement noise level ($\sigma=0.3$), refer to Fig.~\ref{fig:high_meas_grid}.
These results suggest that \Method could be more appropriate for solving challenging inverse problems where measurement noise is prominent, such as low-dose CT reconstruction.

\begin{figure*}[htbp]
    \centering
    \begin{subfigure}[b]{0.49\textwidth}
        \centering
        \includegraphics[width=\linewidth]{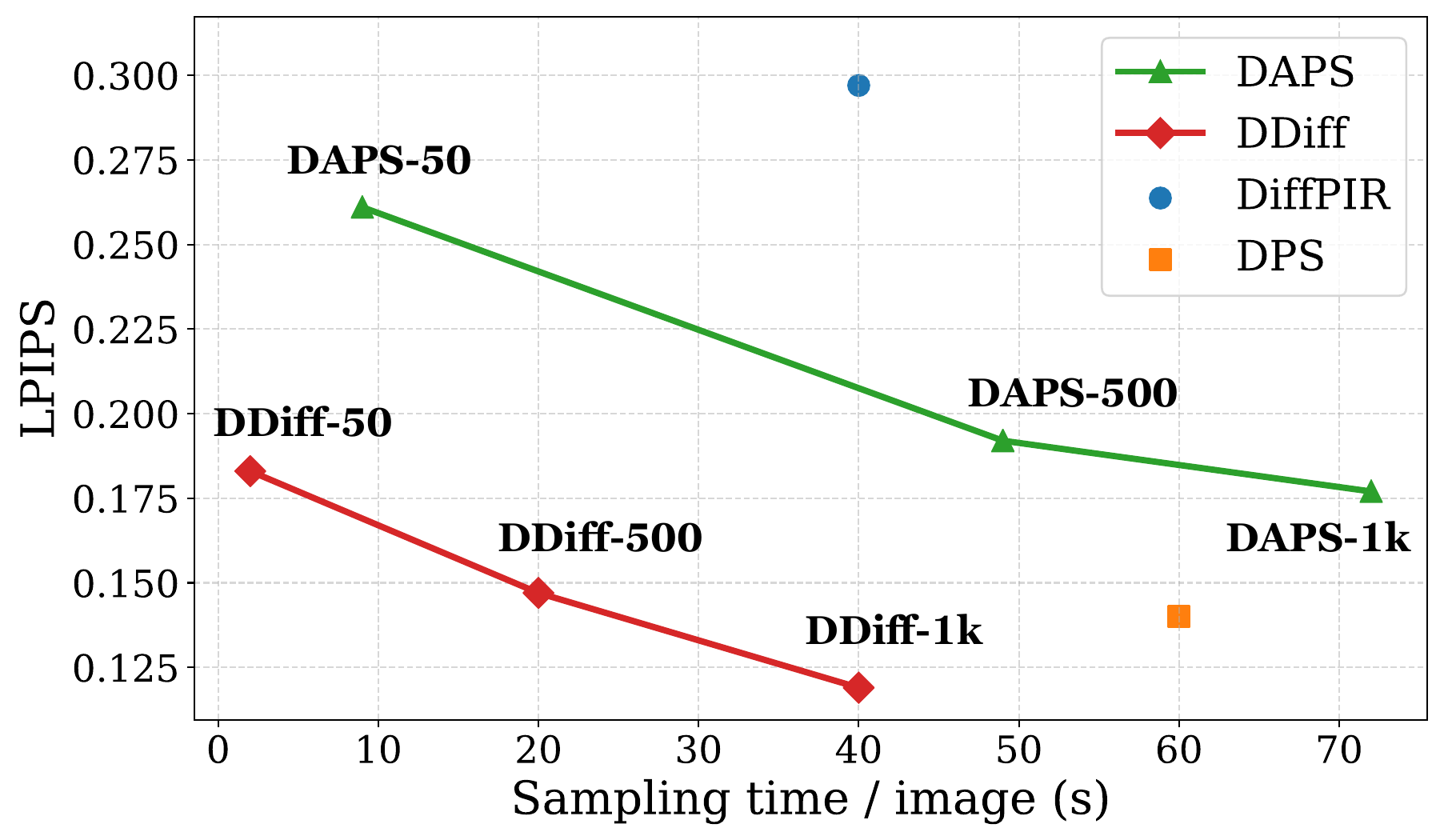}
        \caption{Gaussian deblurring}
        \label{fig:deblur}
    \end{subfigure}
    \hspace{0.1mm}
    \begin{subfigure}[b]{0.49\textwidth}
        \centering
        \includegraphics[width=\linewidth]{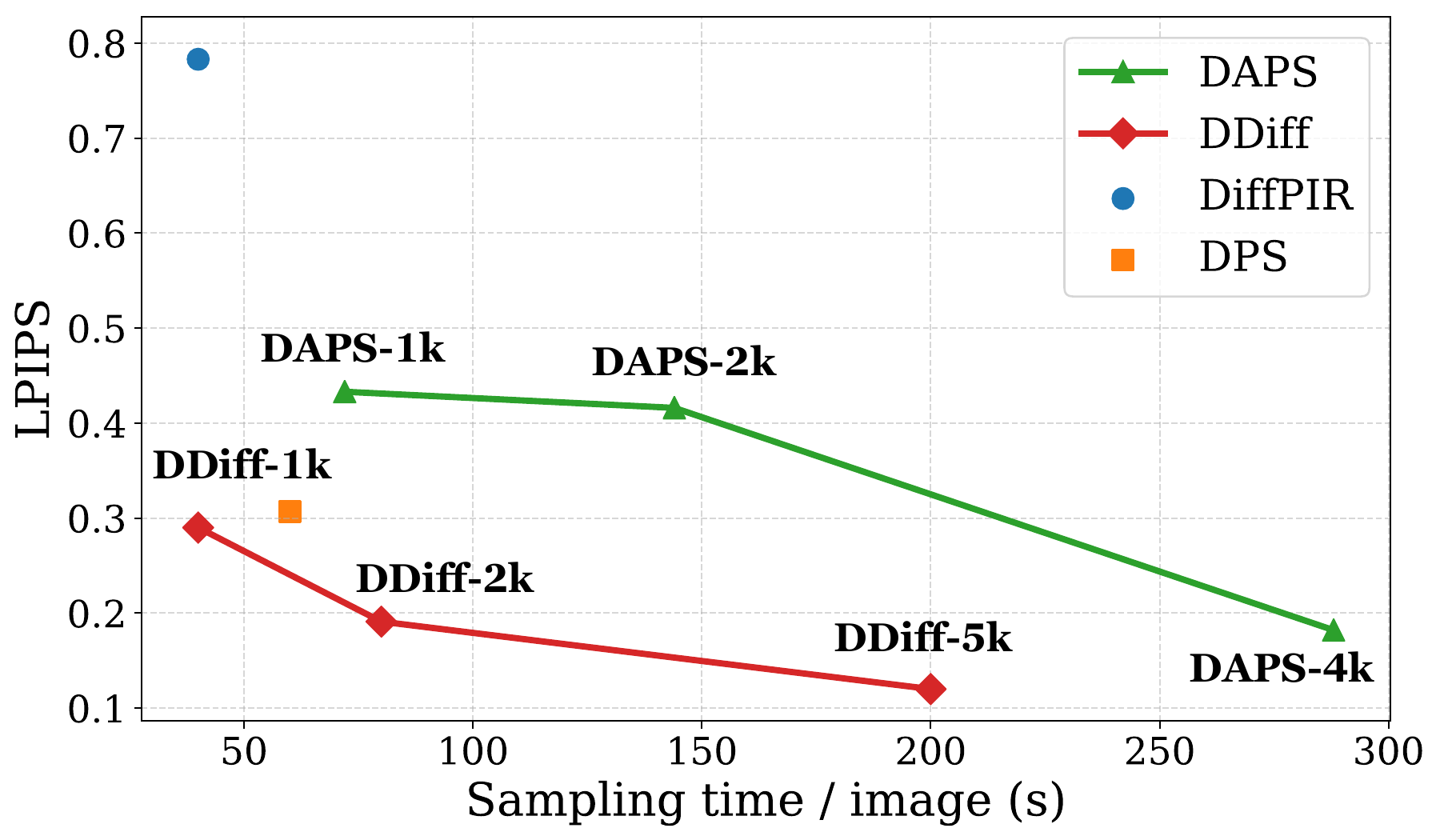}
        \caption{Phase retrieval}
        \label{fig:pr}
    \end{subfigure}
    \caption{\textbf{Evaluation of time efficiency and quality of samples.} The y-axis shows LPIPS value and the x-axis shows the time (in sec.) taken to generate one sample image on a GeForce RTX 2080 Ti 12GB GPU. The number after the method name (500, 2k, etc.) indicates the NFEs. This evaluation uses 100 FFHQ validation images.}
    \label{fig:sample_time}
\end{figure*}

\begin{table*}
  \caption{\textbf{Ablating the dual variable and noising step.} We compare \Method (with noise, with $\mathbf{u}$) to other variants, including Diff-PnP-HQS (no noise, no $\mathbf{u}$), Diff-PnP-ADMM (no noise, with $\mathbf{u}$), and DDiff-HQS (with noise, no $\mathbf{u}$). Full algorithmic descriptions of each variant are provided in Appendix~\ref{appendix:variants}. All methods are compared in a fair manner by using the same DDIM noise schedule. This evaluation uses 10 FFHQ validation images and reports the average metric value. The best and second-best results are distinguished by \textbf{bold} and \underline{underlined} marks, respectively. \Method, with the dual variable and noising step, significantly outperforms other methods.}
  \vspace{5pt}
  \ra{1.2}
  \centering
  \footnotesize
  \begin{adjustbox}{max width=\textwidth}
  \begin{tabular}{l|cccc|cccc|cccc|cccc}
    \toprule
    & \multicolumn{4}{c|}{\textbf{Super Resolution 4×}} & \multicolumn{4}{c|}{\textbf{Inpainting (Box)}} & \multicolumn{4}{c|}{\textbf{Inpainting (Random)}} & \multicolumn{4}{c}{\textbf{Gaussian Deblurring}} \\
    \textbf{Method} & PSNR & SSIM & LPIPS & Res. & PSNR & SSIM & LPIPS & Res. & PSNR & SSIM & LPIPS & Res. & PSNR & SSIM & LPIPS & Res. \\
    & $\uparrow$ & $\uparrow$ & $\downarrow$ & $\downarrow$ & $\uparrow$ & $\uparrow$ & $\downarrow$ & $\downarrow$ & $\uparrow$ & $\uparrow$ & $\downarrow$ & $\downarrow$ & $\uparrow$ & $\uparrow$ & $\downarrow$ & $\downarrow$ \\
    \midrule
    \Method (ours) & \textbf{30.10} & \textbf{0.821} & \textbf{0.199} & \textbf{0.0034} & \textbf{26.01} & \textbf{0.855} & \textbf{0.087} & \textbf{0.0032} & \textbf{33.21} & \textbf{0.883} & \textbf{0.043} & \textbf{0.0028} & \textbf {29.08} & \textbf{0.801} & \textbf{0.112} & \textbf{0.0025} \\
    DDiff-HQS & \underline{25.45} & \underline{0.558} & \underline{0.369} & \underline{0.0057} & \underline{19.01} & \underline{0.566} & \underline{0.221} & \underline{0.0091} & 15.37 & 0.254 & 0.790 & 0.0167 & \underline{26.32} & \underline{0.642} & \underline{0.267} & \underline{0.0031} \\
    Diff-PnP-ADMM & 13.79 & 0.434 & 0.572 & 0.0304 & 12.72 & 0.546 & 0.486 & 0.0298 & \underline{16.79} & \underline{0.540} & \underline{0.402} & \underline{0.0087} & 6.18 & 0.077 & 0.820 & 0.2552 \\
    Diff-PnP-HQS & 14.04 & 0.447 & 0.535 & 0.0287 & 12.61 & 0.557 & 0.476 & 0.0271 & 15.85 & 0.513 & 0.431 & 0.0168 & 5.82 & 0.036 & 0.858 & 0.2812 \\
    \midrule
    & \multicolumn{4}{c|}{\textbf{Motion Deblurring}} & \multicolumn{4}{c|}{\textbf{Phase Retrieval}} & \multicolumn{4}{c|}{\textbf{Nonlinear Deblurring}} & \multicolumn{4}{c}{\textbf{High Dynamic Range}} \\
    \textbf{Method} & PSNR & SSIM & LPIPS & Res. & PSNR & SSIM & LPIPS & Res. & PSNR & SSIM & LPIPS & Res. & PSNR & SSIM & LPIPS & Res. \\
    & $\uparrow$ & $\uparrow$ & $\downarrow$ & $\downarrow$ & $\uparrow$ & $\uparrow$ & $\downarrow$ & $\downarrow$ & $\uparrow$ & $\uparrow$ & $\downarrow$ & $\downarrow$ & $\uparrow$ & $\uparrow$ & $\downarrow$ & $\downarrow$ \\
    \midrule
    \Method (ours) & \textbf{28.14} & \textbf{0.779} & \textbf{0.112} & \textbf{0.0026} & \textbf{30.58} & \textbf{0.834} & \textbf{0.096} & \textbf{0.0028} & \textbf{31.25} & \textbf{0.867} & \textbf{0.111} & \textbf{0.0033} & \textbf{26.95} & \textbf{0.877} & \textbf{0.109} & \textbf{0.0072} \\
    DDiff-HQS & \underline{21.75} & \underline{0.396} & \underline{0.397} & \underline{0.0036} & 12.86 & 0.114 & 0.679 & \underline{0.0057} & \underline{13.51} & \underline{0.186} & \underline{0.543} & \underline{0.1252} & \underline{17.76} & \underline{0.672} & \underline{0.284} & \underline{0.1260} \\
    Diff-PnP-ADMM & 6.35 & 0.089 & 0.810 & 0.1840 & 12.84 & 0.266 & 0.637 & 0.0081 & 5.23 & 0.007 & 0.830 & 0.2463 & 7.85 & 0.126 & 0.875 & 0.4797 \\
    Diff-PnP-HQS & 6.39 & 0.091 & 0.813 & 0.1752 & \underline{13.63} & \underline{0.287} & \underline{0.597} & 0.0077 & 6.02 & 0.086 & 0.827 & 0.2537 & 7.56 & 0.098 & 0.885 & 0.5287 \\
    \bottomrule
  \end{tabular}
  \end{adjustbox}
  \label{tab:three_variants_table}
\end{table*}

\subsection{Ablation Studies}
\paragraph{Evaluation of time efficiency and quality of samples.} One of the most limiting factors of sampling speed for diffusion-based methods is the number of neural function evaluations (NFEs). It measures the number of score model evaluations performed during inference. Therefore, we assess the average sampling time and sample quality of our method and of existing baselines as a function of the number of NFEs in Fig.~\ref{fig:sample_time}. For the same number of NFEs, \Method achieves better perceptual quality and faster sampling speed compared to DAPS. The enhanced computational efficiency is a consequence of two key factors: reduced backpropagation requirements for gradient calculations and the absence of supplementary MCMC procedures within sampling iterations, both of which are present in the DAPS framework.


\paragraph{Effect of dual variable and noising step.} To illustrate the importance of dual variable $\mathbf{u}$ and the additional noising step (Eq.~\ref{eq:forward}) in our algorithm, we conduct an experiment to compare \Method (with noise, with $\mathbf{u}$) to three other variants: Diff-PnP-HQS (no noise, no $\mathbf{u}$), Diff-PnP-ADMM (no noise, with $\mathbf{u}$), and DDiff-HQS (with noise, no $\mathbf{u}$) in Table~\ref{tab:three_variants_table}. Our analysis indicates that the dual variable implementation alone decreases performance, as evidenced by Diff-PnP-ADMM's inferior metrics compared to Diff-PnP-HQS. Without the noising step, the dual variable introduces high-frequency artifacts that compromise the diffusion model's efficacy. However, when incorporated alongside the noising step, the dual variable yields a substantial enhancement in performance, as evidenced by the superior quantitative metrics achieved by \Method compared to DDiff-HQS. Our experimental results confirm that both components\textemdash dual variable and noising step\textemdash must be implemented together, as demonstrated by the significant performance improvement observed when transitioning from Diff-PnP-HQS to \Method.

\section{Conclusion}

In this work, we introduce \Method, a dual-ascent framework for solving diffusion model-based inverse problems that achieves better quantitative and qualitative reconstruction quality than the state of the art, especially at high measurement noise levels. By jointly optimizing in the primal and dual spaces and carefully adapting the ADMM $\mathbf{z}$-update, our approach effectively leverages pretrained diffusion priors to improve perceptual quality, measurement consistency, and runtime. \Method is highly robust to noise and computationally efficient, making it practical for solving large-scale, complex inverse problems. Future work includes extending the framework to higher-dimensional data, such as 3D or video representations, which presents additional challenges and opportunities.

{
    \small
    \bibliographystyle{ieeenat_fullname}
    \bibliography{main}
}


\clearpage
\onecolumn                
\setcounter{page}{1}

\begin{center}
  {\Large \bfseries Dual Ascent Diffusion for Inverse Problems \\[10pt]} 
  {\large Supplementary Material \\[4pt]}
\end{center}
\renewcommand{\thesection}{\Alph{section}}
\renewcommand{\thesubsection}{\thesection.\arabic{subsection}}
\setcounter{section}{0}  

\appendix
\section{Convergence Analysis of DDiff}
\label{appendix:convergence}
\begin{multicols}{2}
\raggedcolumns

\begingroup
\section*{Notations}

\begin{itemize}
    \item $X \subset \mathbb{R}^n$.
    \item For $v\in\mathbb{R}^n$, $\|v\| := \|v\|_2/\sqrt{n}$.
    \item (Forward operator) $A:\mathbb{R}^n\mapsto\mathbb{R}^m$.
    \item (Measurement noise level) $\sigma\in\mathbb{R}$.
    \item (Measurement) $y\in\mathbb{R}^m$.
    \item (Data fidelity term) $f(x) := -\frac{1}{2\sigma^2}\,\|y-A(x)\|_2^2$.
    \item (Diffusion times) $\{t_k\}_{k\geq 0}\in\mathbb{R}_+^\mathbb{N}$.
    \item (Schedule parameters) $\forall k\in\mathbb{N}, \bar{\alpha}_{t_k}\in(0,1], \sigma_{t_k}\in\mathbb{R}_+$.
    \item (Score network) $s_\theta:\mathbb{R}^n\times\mathbb{R}_+\mapsto\mathbb{R}^n$
    \item (Noise threshold) $t_{\text{TH}}\in\mathbb{R}_+$
    \item (Step sizes) $\{\gamma_k\}_{k\geq 0}\in\mathbb{R}_+^\mathbb{N}$.
\end{itemize}
\columnbreak
\section*{Assumptions}

\begin{enumerate}[(i)]
    \item $X$ is compact (e.g., $X=[-1,1]^n$).
    \item $\lim_{k\to\infty} t_k=0$.
    \item $\lim_{k\to\infty} \bar{\alpha}_{t_k}=1$.
    \item $\lim_{k\to\infty} \sigma_{t_k}=0$.
    \item $s_\theta$ is bounded on $X$, i.e. $$\exists S\in\mathbb{R}_+~|~\forall(\tilde{x},t)\in X\times\mathbb{R}_+,\Vert s_\theta(\tilde{x},t)\Vert\leq S.$$
    This is standard in practice, e.g., with ReLU/Tanh final layers on normalized inputs or via runtime clipping to enforce bounded outputs.
    \item $\lim_{k\to\infty} \gamma_k=0$.
    \item $\sum_{k=0}^\infty \gamma_k<\infty$
    \item $\sum_{k=0}^\infty\sigma_{t_k} < \infty$
    \item $\sum_{k=0}^\infty\sqrt{1-\bar{\alpha}_{t_k}} < \infty$
\end{enumerate}
\end{multicols}
\endgroup

\section*{DDiff Recursive Process}




\subsection*{Initialization}

$(x_0,z_0,u_0)\in X\times X\times \mathbb{R}^n$ (e.g., $x_0=z_0$ and $u_0=0$) and $\tilde{x}_{t_0}\sim \mathcal{N}(0,I)$.

\subsection*{Recurrence Relation}

For $k\in\mathbb{N}$,
\[
\begin{aligned}
&\text{z-update (Tweedie):} &&
z_{k+1} \;=\; \frac{1}{\sqrt{\bar\alpha_{t_k}}}\Big(\tilde{x}_{t_k} + (1-\bar\alpha_{t_k})\, s_\theta(\tilde{x}_{t_k},t_k)\Big),\\[4pt]
&\text{x-update (linearized ADMM):} &&
x_{k+1} \;=\; v_k + \gamma_k \nabla f(v_k), \quad v_k := z_{k+1} - u_k,\\[4pt]
&\text{dual update:} &&
u_{k+1} \;=\; u_k + x_{k+1} - z_{k+1}.
\end{aligned}
\]

DDIM-style latent reverse step anchored at $(x_{k+1},u_k)$:
\[
\hat\varepsilon_{t_k} \;=\; \frac{\tilde{x}_{t_k}-\sqrt{\bar\alpha_{t_k}}\,x_{k+1}}{\sqrt{1-\bar\alpha_{t_k}}},\qquad
\varepsilon_{t_k}\sim
\begin{cases}
\mathcal N(0,I), & t_k>t_{\mathrm{TH}},\\
0, & t_k\le t_{\mathrm{TH}}.
\end{cases}
\]

\begin{align}
    \tilde{x}_{t_{k+1}}
&=\sqrt{\bar\alpha_{t_{k+1}}}\,x_{k+1} + \sqrt{\bar\alpha_{t_{k+1}}}\,u_k
+ \sqrt{1-\bar\alpha_{t_{k+1}}-\sigma_{t_k}^2}\,\hat\varepsilon_{t_k}
+ \sigma_{t_k}\,\varepsilon_{t_k}\\
&=\sqrt{\bar\alpha_{t_{k+1}}}\,(x_{k+1} + u_k)
+ \sqrt{1-\bar\alpha_{t_{k+1}}-\sigma_{t_k}^2}\,\hat\varepsilon_{t_k}
+ \sigma_{t_k}\,\varepsilon_{t_k}
\end{align}
The ``signal part'' of $\Tilde{x}_{t_{k+1}}$ in the reverse step is $\sqrt{\bar\alpha_{t_{k+1}}}\,w_k:=\sqrt{\bar\alpha_{t_{k+1}}}\,(x_{k+1}+u_k)$. We say the latent $\Tilde{x}_{t_k}$ is ``anchored'' at $w_k$ scaled by $\sqrt{\bar\alpha_{t_{k+1}}}$. The subsequent Tweedie z-update $z_{k+2}$ is then a denoiser applied \emph{around} this same anchor, so $z_{k+2}$ stays near $w_k$ when $t_k$ is small and the schedule is in its deterministic tail, i.e., $t_k\le t_{\mathrm{TH}}$. This “anchoring” reproduces the ADMM pattern where the denoiser acts on $x_{k+1}+u_k$ rather than on $x_{k+1}$ alone.

\vspace{0.5em}
\noindent In this section, we aim to establish the fixed-point convergence of DDiff, following the key lemmas presented below.

\section*{Lemmas}

\begin{lemma}[Bounded diffusion denoiser with vanishing strength]\label{lem:bounded-denoiser}
\[
\forall k\in\mathbb{N},
\|z_{k+1}-w_{k-1}\| \;\le\; C_d\,\sigma_{\mathrm{eff},t_k},
\]
where
\[
w_{k-1}:=x_k+u_{k-1}
\]
and
\[
\sigma_{\mathrm{eff},t_k}:=\sqrt{1-\bar\alpha_{t_k}} \;+\; \sigma_{t_{k-1}} \;+\; (1-\bar\alpha_{t_k})
\]
for some constant $C_d$.

\begin{proof}
Write the z-update and subtract the anchor to see the deviation:
\[
z_{k+1}-w_{k-1} \;=\; \frac{1}{\sqrt{\bar\alpha_{t_k}}}\Big(\Tilde{x}_{t_k} - \sqrt{\bar\alpha_{t_k}}w_{k-1}\Big)
\;+\; \frac{1-\bar\alpha_{t_k}}{\sqrt{\bar\alpha_{t_k}}}\, s_\theta(\Tilde{x}_{t_k},t_k).
\]
The reverse step with anchor $w_{k-1}$ yields
\[
\Tilde{x}_{t_k} \;=\; \sqrt{\bar\alpha_{t_k}}\,w_{k-1} \;+\; \delta_{t_k},\qquad
\delta_{t_k} := \sqrt{1-\bar\alpha_{t_k}-\sigma_{t_{k-1}}^2}\,\hat\varepsilon_{t_{k-1}} + \sigma_{t_{k-1}} \varepsilon_{t_{k-1}},
\]
where $\hat\varepsilon_{t_k}$ is the network-induced direction with $\|\hat\varepsilon_{t_k}\|/\sqrt{n}\lesssim 1$ on bounded $X$ and $\|\varepsilon_{t_k}\|/\sqrt{n}\approx 1$ for large $n$ almost surely. Hence
\[
\frac{1}{\sqrt{n}}\|\Tilde{x}_{t_k}-\sqrt{\bar\alpha_{t_k}}w_{k-1}\| \;=\; \frac{1}{\sqrt{n}}\|\delta_{t_k}\| \;\le\; c_1\sqrt{1-\bar\alpha_{t_k}} + c_2 \sigma_{t_{k-1}}
\]
for some constants $c_1,c_2 > 0$ and given $\sigma_{t_k}\ge0$. 
Using $\|s_\theta(x_{t_k},t_k)\|\le S_{t_k}$ and boundedness of $X$,
\[
\frac{1}{\sqrt{n}}\|z_{k+1}-w_{k-1}\|
\;\le\;
\frac{c_1}{\sqrt{\bar\alpha_{t_k}}}\sqrt{1-\bar\alpha_{t_k}}
+ \frac{c_2}{\sqrt{\bar\alpha_{t_k}}}\sigma_{t_{k-1}}
+ \frac{1-\bar\alpha_{t_k}}{\sqrt{\bar\alpha_{t_k}}}\,\frac{S_{t_k}}{\sqrt{n}}
\;\le\;
C_d\big(\sqrt{1-\bar\alpha_{t_k}} + \sigma_{t_{k-1}} + (1-\bar\alpha_{t_k})\big),
\]
with $C_d$ absorbing constants. For a fixed, finite schedule $\bar\alpha_{\min}:=\min_{t_k} \bar\alpha_{t_k}>0$ and finite $S_{t_k}$, $\sup_{k} \frac{S_{t_k}}{\sqrt{n}\sqrt{\bar\alpha_{t_k}}}<\infty$; hence both
$1/\sqrt{\bar\alpha_{t_k}}$ and $S_{t_k}$ are uniformly bounded and can be absorbed into $C_d$.
Concretely, one may take
\[
C_d \;\ge\; \max\!\left\{
\frac{\max\{c_1,c_2\}}{\sqrt{\bar\alpha_{\min}}},
\;\;
\sup_{k} \frac{S_{t_k}}{\sqrt{n}\sqrt{\bar\alpha_{t_k}}}
\right\}.
\] Define effective denoiser strength 
\[
\sigma_{\mathrm{eff},t_k} \;:=\; \sqrt{1-\bar\alpha_{t_k}} \;+\; \sigma_{t_{k-1}} \;+\; (1-\bar\alpha_{t_k}).
\]
As $t_k\rightarrow0$, $1-\bar\alpha_{t_k} \rightarrow 0$ and $\sigma_{t_k} \rightarrow0$ (and at the deterministic tail, we have exactly $\sigma_{t_k}=0$), so $\sigma_{\mathrm{eff},t_k}\rightarrow0$.
\end{proof}

\begin{remark}[Nonconvex prior is compatible with Lemma \ref{lem:bounded-denoiser}]
Lemma \ref{lem:bounded-denoiser} does \emph{not} require convexity of a prior; it only requires that the z-map be close to the identity when the diffusion noise and Tweedie correction are small (i.e., denoiser $\mathcal{D}_\sigma \rightarrow \mathcal{I}$ as $\sigma\rightarrow0$ \cite{chan2016plug}). Proof of Lemma \ref{lem:bounded-denoiser} delivers this by showing $z_{k+1}\approx w_{k-1}$ with an error bounded by $\sigma_{\mathrm{eff},t_k}$ as $t_k\rightarrow0$, even if the learned score (hence implicit prior) is nonconvex.
\end{remark}

\end{lemma}

\begin{lemma}[Bounded gradient of the data term]\label{lem:bounded-grad}
Let $f(x):=-\|y-A(x)\|_2^2/(2\sigma^2)$. There exists $L<\infty$ such that
\[
\sup_{x\in X}\,\|\nabla f(x)\| \;\le\; L.
\]
\end{lemma}

\begin{proof}
Assume that $A : X \to \mathbb{R}^m$ is continuously differentiable on the compact set 
$X = [-1,1]^n$, so its Jacobian $J_A(x)$ exists and is continuous. 
Define 
\[
f(x) = -\frac{1}{2\sigma^2}\|y - A(x)\|_2^2.
\]

By the chain rule,
\[
\nabla f(x) = -\frac{1}{\sigma^2} J_A(x)^\top (y - A(x)).
\]
Hence,
\[
\|\nabla f(x)\| 
\;\le\; \frac{1}{\sigma^2} \|J_A(x)^\top\| \, \|y - A(x)\|
\;=\; \frac{1}{\sigma^2} \|J_A(x)\|_{\mathrm{op}} \, \|y - A(x)\|,
\]
where $\|\cdot\|_{\mathrm{op}}$ denotes the operator (spectral) norm of a matrix.

Since $J_A(x)$ and $A(x)$ are continuous on the compact set $X$, 
the functions $x \mapsto \|J_A(x)\|_{\mathrm{op}}$ and $x \mapsto \|A(x)\|$ 
are continuous and therefore bounded by the boundedness theorem. 
Let
\[
M_J := \sup_{x \in X} \|J_A(x)\|_{\mathrm{op}} < \infty, 
\qquad
M_A := \sup_{x \in X} \|A(x)\| < \infty.
\]
Then, for all $x \in X$,
\[
\|\nabla f(x)\| 
\;\le\; \frac{1}{\sigma^2} M_J (\|y\| + \|A(x)\|)
\;\le\; \frac{1}{\sigma^2} M_J (\|y\| + M_A).
\]
Define
\[
L := \frac{1}{\sigma^2} M_J (\|y\| + M_A) < \infty.
\]
It follows that
\[
\sup_{x \in X} \|\nabla f(x)\| \;\le\; L,
\]
which proves the lemma.
\end{proof}

\begin{lemma}[Closed-form dual update and vanishing behavior under bounded gradients]\label{lem:dual-vanishing}
Suppose the iterates $\{v_k\}$ lie in $X$, then
\[
u_{k+1} \;=\; \,\gamma_k\,\nabla f(v_k),
\]
and, using Lemma~\ref{lem:bounded-grad},
\[
\|u_{k+1}\| \;=\; \gamma_k \,\|\nabla f(v_k)\|
\;\le\; \gamma_k\,L \;\xrightarrow[k\to\infty]{}\; 0.
\]

\proof
By definition,
\[
u_{k+1}=u_k+x_{k+1}-z_{k+1}.
\]
Using the $x$-update and $v_k:=z_{k+1}-u_k$,
\[
x_{k+1}=v_k+\gamma_k\nabla f(v_k)
\quad\Longrightarrow\quad
u_{k+1}=u_k+(v_k+\gamma_k\nabla f(v_k))-z_{k+1}
= \gamma_k\nabla f(v_k).
\]
Taking norms and applying Lemma~\ref{lem:bounded-grad} yields the claim.
\qed
\end{lemma}

\begin{lemma}[u-step increment]\label{lem:u-increment}
Under Lemma~\ref{lem:dual-vanishing},
\[
\|u_{k+1}-u_k\| \;=\; \|x_{k+1}-z_{k+1}\| \;\le\; L(\,\gamma_k\,+\,\gamma_{k-1}).
\]
\end{lemma}

\begin{proof}
$\|u_{k+1}-u_k\|\le \|u_k\|+\|u_{k+1}\|\le L(\,\gamma_k\,+\,\gamma_{k-1})$, where we used $\|u_k\|\le L\,\gamma_{k-1}$ inductively.
\end{proof}

\begin{lemma}[x-step increment]\label{lem:x-increment}
With Lemma \ref{lem:bounded-denoiser} and Lemma~\ref{lem:u-increment}, we obtain for $k\ge 2$: 
\[
\|x_{k+1}-x_k\| \;\le\; C_d\,\sigma_{\mathrm{eff},t_k} \,+\, L(\,\gamma_k +\gamma_{k-1}+\gamma_{k-2}).
\]
\end{lemma}

\begin{proof}
Let $v_k=z_{k+1}-u_k$ and $x_{k+1}=v_k+\gamma_k\nabla f(v_k)$ with $f(v)=-\|y-A(v)\|_2^2/2\sigma^2$. Add and subtract $v_k$, we obtain $\|x_{k+1}-x_k\|=\|x_{k+1}-x_k+v_k-v_k\|\le \|x_{k+1}-v_k\| + \|x_k-v_k\|$. We know that $\|x_{k+1}-v_k\|=\gamma_k\|\nabla f(v_k)\|\le \gamma_k L$. To bound $\|x_k-v_k\|$, we decompose $v_k-x_k=(z_{k+1}-(x_k+u_{k-1}))+(u_{k-1}-u_k)$ and then apply Lemma \ref{lem:bounded-denoiser} with Lemma~\ref{lem:u-increment}, 
$$
\begin{aligned}
\|x_k - v_k\|
&\le \|z_{k+1} - (x_k + u_{k-1})\| + \|u_k - u_{k-1}\| \\
&= \|z_{k+1} - w_{k-1}\| + \|u_k - u_{k-1}\|
\\
&\le C_d\,\sigma_{\mathrm{eff},t_k} \,+\, L(\,\gamma_{k-1}\,+\,\gamma_{k-2})
\end{aligned}
$$
Combining the bounds yields $\|x_{k+1}-x_k\| \le \|x_{k+1}-v_k\| + \|x_k-v_k\| \le C_d\,\sigma_{\mathrm{eff},t_k} \,+\, L(\,\gamma_k +\gamma_{k-1}+\gamma_{k-2})$.

\end{proof}

\begin{lemma}[z-step increment]\label{lem:z-increment}
Let $t_k$ be the diffusion index used to form $z_{k+1}$ (so $z_k$ was formed at $t_{k-1}$). Under Lemma~\ref{lem:bounded-denoiser}--\ref{lem:x-increment}, for $k\ge 3$,
\[
\|z_{k+1}-z_k\|
\;\le\; C_d\,\sigma_{\mathrm{eff},t_k} \;+\; 2C_d\,\sigma_{\mathrm{eff},t_{k-1}}
\;+\; L\big(\gamma_{k-1}+2\gamma_{k-2}+2\gamma_{k-3}\big).
\]
\end{lemma}

\begin{proof}
Introduce the anchors $w_{k-1}:=x_k+u_{k-1}$ and $w_{k-2}:=x_{k-1}+u_{k-2}$. By the triangle inequality,
\[
\|z_{k+1}-z_k\|
\;\le\; \|z_{k+1}-w_{k-1}\| \;+\; \|w_{k-1}-w_{k-2}\| \;+\; \|w_{k-2}-z_k\|.
\]
By Lemma~\ref{lem:bounded-denoiser},
\[
\|z_{k+1}-w_{k-1}\| \le C_d\,\sigma_{\mathrm{eff},t_k},
\qquad
\|z_k-w_{k-2}\|\le C_d\,\sigma_{\mathrm{eff},t_{k-1}}.
\]
For the anchor increment,
\[
\|w_{k-1}-w_{k-2}\|
\;\le\; \|x_k-x_{k-1}\| + \|u_{k-1}-u_{k-2}\|.
\]
Apply Lemma~\ref{lem:x-increment} at index $k-1$ and Lemma~\ref{lem:u-increment} at index $k-2$:
\[
\|x_k-x_{k-1}\| \;\le\; C_d\,\sigma_{\mathrm{eff},t_{k-1}} + L(\gamma_{k-1}+\gamma_{k-2}+\gamma_{k-3}),
\qquad
\|u_{k-1}-u_{k-2}\| \;\le\; L(\gamma_{k-2}+\gamma_{k-3}).
\]
Combining the bounds yields
\[
\|w_{k-1}-w_{k-2}\|
\;\le\; C_d\,\sigma_{\mathrm{eff},t_{k-1}}
\;+\; L\big(\gamma_{k-1}+2\gamma_{k-2}+2\gamma_{k-3}\big).
\]
Returning to the decomposition, we obtain
\[
\|z_{k+1}-z_k\|
\;\le\; C_d\,\sigma_{\mathrm{eff},t_k} + C_d\,\sigma_{\mathrm{eff},t_{k-1}}
\;+\; \Big[C_d\,\sigma_{\mathrm{eff},t_{k-1}}
\;+\; L\big(\gamma_{k-1}+2\gamma_{k-2}+2\gamma_{k-3}\big)\Big],
\]
which simplifies to the stated inequality.
\end{proof}

\section*{Target Theorem}

\begin{theorem}[Fixed-point Convergence of DDiff]\label{thm:fp-ddiff}
Let $\{(x_k,z_k,u_k)\}_{k\ge 0}$ be generated by the DDiff updates above.
Then,
the sequences $\{x_k\}$, $\{z_k\}$ and $\{u_k\}$ are Cauchy sequences. In particular,
\[
\|x_{k+1}-x_k\|+\|z_{k+1}-z_k\|+\|u_{k+1}-u_k\| \xrightarrow[k\to\infty]{} 0,
\]
and $(x_k,z_k,u_k)\to (x^\star,z^\star,u^\star)$.
\end{theorem}

\begin{proof}

\textbf{Step 1} (increment bounds). Lemma~\ref{lem:u-increment} gives
\[
\|u_{k+1}-u_k\| \;\le\; L(\,\gamma_k\,+\,\gamma_{k-1}).
\]

Lemma~\ref{lem:x-increment} gives
\[
\|x_{k+1}-x_k\| \;\le\; C_d\,\sigma_{\mathrm{eff},t_k} \,+\, L(\,\gamma_k +\gamma_{k-1}+\gamma_{k-2}).
\]

Lemma~\ref{lem:z-increment} gives
\[
\|z_{k+1}-z_k\| \;\le\; C_d\,\sigma_{\mathrm{eff},t_k} \;+\; 2C_d\,\sigma_{\mathrm{eff},t_{k-1}}
\;+\; L\big(\gamma_{k-1}+2\gamma_{k-2}+2\gamma_{k-3}\big).
\]

\textbf{Step 2} (summability). Summing the three bounds,
\begin{align}
\|x_{k+1}-x_k\|+\|z_{k+1}-z_k\|+\|u_{k+1}-u_k\|
&\le 2C_d\,\sigma_{\mathrm{eff},t_k} + 2C_d\,\sigma_{\mathrm{eff},t_{k-1}}\label{eq:first}\\
&\quad + L\Big(2\gamma_k + 3\gamma_{k-1} + 3\gamma_{k-2} + 2\gamma_{k-3}\Big)\notag\\
&\le 2C_d\big(\sigma_{\mathrm{eff},t_k}+\sigma_{\mathrm{eff},t_{k-1}}\big)
+ 3L\sum_{j=0}^3 \gamma_{k-j}. \label{eq:last}
\end{align}

Following Lemma \ref{lem:bounded-denoiser} and the Assumptions, the right-hand side of \eqref{eq:last} is summable over $k$ and tends to $0$ as $k\to\infty$:
\[
\sum_{k=0}^\infty \big(\|x_{k+1}-x_k\|+\|z_{k+1}-z_k\|+\|u_{k+1}-u_k\|\big) < \infty,
\quad
\|x_{k+1}-x_k\|+\|z_{k+1}-z_k\|+\|u_{k+1}-u_k\| \to 0.
\]

Absolute summability implies that $\{x_k\}$, $\{z_k\}$, and $\{u_k\}$ are Cauchy, hence converge in $\mathbb{R}^{3n}$ to $(x^\star,z^\star,u^\star)$.

By Lemma~\ref{lem:dual-vanishing}, we have 
\[
u_{k+1} = \gamma_k \nabla f(v_k), \quad v_k := z_{k+1} - u_k.
\]
Since $\|\nabla f(v_k)\|$ is bounded and $\gamma_k \to 0$, it follows that $u_{k+1} \to 0$, hence $u^\star = 0$. Moreover,
\[
x_{k+1} - z_{k+1} = u_{k+1} - u_k \to 0,
\]
so $x^\star = z^\star$.

Taking $k \to \infty$ in the DDiff updates and noting that $\sigma_{\mathrm{eff},t_k} \to 0$, $\gamma_k \to 0$, and $(x_k, z_k, u_k) \to (x^\star, z^\star, 0)$, we obtain:

\begin{itemize}
    \item By Lemma~\ref{lem:bounded-denoiser}, $z_{k+1} - (x_k + u_{k-1}) \to 0$;
    \item From the $x$-update, $x_{k+1} - (z_{k+1} - u_k) \to 0$;
    \item The dual update $u_{k+1} = u_k + x_{k+1} - z_{k+1}$ holds by definition.
\end{itemize}

In the limit, these relations imply
\[
z^\star = x^\star, \quad x^\star = z^\star, \quad \text{and} \quad u^\star = 0,
\]
confirming that $(x^\star, z^\star, u^\star) = (x^\star, x^\star, 0)$ is a fixed point of the asymptotic DDiff updates.
\end{proof}

\begin{remark}[Finite iterations, step-size policy, and ADMM--diffusion coupling]
We run a finite number of iterations $k=0,\dots,K$, setting $K=T$ equal to the number of diffusion time steps. Each diffusion reverse step $(t_k \!\to\! t_{k-1})$ is paired with one ADMM update, 
so that the dual variable $u_k$ is updated in synchrony with the denoising process. The convergence analysis assumes a vanishing step-size schedule \((\gamma_k \to 0)\). In practice, we adopt a nonincreasing, step-down policy for \(\gamma_k\) to control measurement update magnitudes; the exact schedule and hyperparameters are provided in Sec.~\ref{supp:hyperparam}. Empirically, as \(\gamma_k\) becomes sufficiently small across iterations, the iterates are stable and reconstruction quality improves, as evidenced by lower LPIPS and higher SSIM (often with competitive PSNR).
\end{remark}

\section{Residual Metric}
\label{appendix:residual}
We report the residual 
\[
r(x) = \|y - \mathcal{A}(x)\|_2^2 - \sigma^2,
\]
which measures the deviation of the reconstruction from the noisy measurement after accounting for the expected noise energy $\sigma^2=\mathbb{E}\big[\|\varepsilon\|_2^2\big]$ under the forward model $y=\mathcal{A}(x)+\varepsilon$. 
An ideal posterior sample $\tilde{x}\!\sim\!p(x|y)$ satisfies $\mathbb{E}_{x\sim p(x),y\sim p(y|x),\tilde{x}\sim p(x|y)}[r(\tilde{x})]\!=\!0$. 
Hence, $r(x)$ quantifies \emph{data consistency}: 
small positive values indicate faithful reconstructions that match the measurement statistics, 
whereas excessively negative values (i.e., $\|y - \mathcal{A}(x)\|_2^2 \!\ll\! \sigma^2$) 
correspond to over-smoothed or overly noise-fitting solutions, 
and large positive values indicate hallucination or loss of fidelity to $y$. 

We therefore interpret residual magnitudes near zero as optimal, 
while deviations in either direction (too small or too large) 
signal a mismatch between data fidelity and prior regularization. 
Throughout, we use the residual only as a diagnostic for data consistency 
and report it alongside perceptual and distortion metrics (PSNR, SSIM, LPIPS).

\newpage
\section{DDiff with Latent Diffusion Models}
\label{sec:latent_ddiff_supp}

Pretrained diffusion models can operate in the latent space of an autoencoder instead of directly in the pixel space to reduce dimensionality and improve computational efficiency~\cite{ldm}. 
An encoder $\mathcal{E}:\mathbb{R}^{H\times W\times C}\!\to\!\mathbb{R}^{d}$ maps an image $\mathbf{x}$ to a compact latent representation $\mathbf{z}=\mathcal{E}(\mathbf{x})$, and a decoder $\mathcal{D}:\mathbb{R}^{d}\!\to\!\mathbb{R}^{H\times W\times C}$ reconstructs an image $\tilde{\mathbf{x}}=\mathcal{D}(\mathbf{z})$. 
Diffusion training and denoising are then performed in the latent space $\mathbf{z}$ using a latent score network $\mathbf{s}_\theta(\mathbf{z},t)$, which estimates $\nabla_{\mathbf{z}}\log p_t(\mathbf{z})$. 
Given a pretrained LDM, our goal is to adapt \Method to operate with such encoder–decoder architectures.

\begin{algorithm}[htbp]
\caption{LatentDDiff (latent-space DDiff with an encoder--decoder LDM)}
\label{alg:latent_ddiff}
\begin{algorithmic}[1]
\REQUIRE $T$; forward operator $\mathcal{A}(\cdot)$; schedules $\{\sigma_t\}_{t=1}^{T}$, $\{\bar{\alpha}_t\}_{t=1}^{T}$; latent score $\mathbf{s}_\theta$; measurements $\mathbf{y}$; step sizes $\{\gamma_t\}_{t=1}^T$; noise threshold $t_0$; encoder $\mathcal{E}$, decoder $\mathcal{D}$; \textsc{Mode} $\in\{\textsc{Latent-DC},\textsc{Pixel-DC}\}$
\STATE Initialize latent variable $\mathbf{z}_T \sim \mathcal{N}(\mathbf{0},\mathbf{I})$, dual variable $\mathbf{u}=\mathbf{0}$.
\FOR{$t = T-1$ $\textbf{to}$ $0$}
    \STATE $\tilde{\mathbf{z}} \leftarrow \frac{1}{\sqrt{\bar{\alpha}_t}}\!\left(\mathbf{z}_t + (1-\bar{\alpha}_t)\,\mathbf{s}_\theta(\mathbf{z}_t,t)\right)$ \hfill {\color{gray}$\triangleright$ Denoising in latent space}
    \STATE \textit{Data-consistency update (choose one by \textsc{Mode})}
    \IF{\textsc{Mode} $=$ \textsc{Latent-DC}}
        \STATE $\hat{\mathbf{x}} \leftarrow \tilde{\mathbf{z}} - \mathbf{u} - \gamma_t\,\nabla_{\mathbf{v}=\tilde{\mathbf{z}}-\mathbf{u}} \,\big\|\mathbf{y}-\mathcal{A}(\mathcal{D}(\mathbf{v}))\big\|_2^2$ \hfill {\color{gray}$\triangleright$ Gradient flows through $\mathcal{D}$: $\nabla_{\mathbf{v}}\|\cdot\|^2$ involves $J_{\mathcal{D}}(\mathbf{v})$}
        \STATE $\mathbf{x} \leftarrow \mathcal{E}(\mathcal{D}(\hat{\mathbf{x}}))$ \hfill {\color{gray}$\triangleright$ Re-Encode}
    \ELSE 
        \STATE $\tilde{\mathbf{x}} \leftarrow \mathcal{D}(\tilde{\mathbf{z}})$ \hfill {\color{gray}$\triangleright$ \textsc{Pixel-DC}}
        \STATE $\hat{\mathbf{x}} \leftarrow \tilde{\mathbf{x}} - \mathbf{u} - \gamma_t\,\nabla_{\mathbf{v}=\tilde{\mathbf{x}}-\mathbf{u}} \,\big\|\mathbf{y}-\mathcal{A}(\mathbf{v})\big\|_2^2$
        \STATE $\mathbf{x} \leftarrow \mathcal{E}(\hat{\mathbf{x}})$ \hfill {\color{gray}$\triangleright$ Encode to latent}
    \ENDIF
    \STATE $\hat{\epsilon}\leftarrow \frac{1}{\sqrt{1-\bar{\alpha}_t}}\!\left(\mathbf{z}_t - \sqrt{\bar{\alpha}_t}\,\mathbf{x}\right)$
    \STATE $\epsilon \sim \mathcal{N}(\mathbf{0},\mathbf{I}) \;\textbf{if}\; t>t_0 \;\textbf{else}\; \epsilon=0$
    \STATE $\mathbf{z}_{t-1} \leftarrow \sqrt{\bar{\alpha}_{t-1}}\mathbf{x}
            + \sqrt{1-\bar{\alpha}_{t-1}-\sigma_t^2}\,\hat{\epsilon}
            + \sigma_t\,\epsilon
            + \sqrt{\bar{\alpha}_{t-1}}\,\mathbf{u}$ \hfill {\color{gray}$\triangleright$ Reverse diffusion in latent space}
    \STATE $\mathbf{u} \leftarrow \mathbf{u} + \mathbf{x} - \tilde{\mathbf{z}}$ \hfill {\color{gray}$\triangleright$ Dual update}
\ENDFOR
\STATE \textbf{return} $\mathbf{x}_0 \leftarrow \mathcal{D}(\mathbf{z}_0)$
\end{algorithmic}
\end{algorithm}

\paragraph{Latent-space diffusion updates.} 
Analogous to the pixel-space formulation, each DDiff iteration in the latent space performs (i) a denoising step using the latent score and (ii) a data-consistency (DC) correction. 
Let $\mathbf{z}_t$ denote the latent variable at timestep $t$ and $\bar{\alpha}_t$ the corresponding noise schedule. 
The denoising prediction is given by
\begin{equation}
\tilde{\mathbf{z}} = \frac{1}{\sqrt{\bar{\alpha}_t}}\!\left(\mathbf{z}_t + (1-\bar{\alpha}_t)\,\mathbf{s}_\theta(\mathbf{z}_t,t)\right),
\label{eq:latent_denoise}
\end{equation}
after which a DC update is applied before the reverse diffusion step.

\paragraph{Data-consistency in latent diffusion.}
Incorporating measurement consistency within latent diffusion can be done in two ways, depending on where the update is applied:

\paragraph{(1) \textsc{Latent-DC}: data consistency in latent space.}
This approach enforces the measurement constraint directly in the latent space by minimizing the following objective: 
\begin{equation}
\hat{\mathbf{x}} 
\leftarrow \operatorname*{argmin}_{\hat{\mathbf{x}}}
\frac{1}{2\sigma^2}\big\|\mathbf{y}-\mathcal{A}(\mathcal{D}(\hat{\mathbf{x}}))\big\|_2^2
+ \frac{\rho}{2}\big\|\hat{\mathbf{x}}-\tilde{\mathbf{z}}+\mathbf{u}\big\|_2^2,
\label{eq:dc_objective}
\end{equation}
where the first term enforces fidelity to the measurements and the second term couples the current estimate $\hat{\mathbf{x}}$ with the diffusion variable $\tilde{\mathbf{z}}$ and dual variable $\mathbf{u}$. In practice, rather than solving~\eqref{eq:dc_objective} exactly, a single gradient step is performed to approximate the proximal update, which is well established in the plug-and-play literature as an efficient first-order method for nonlinear forward operators~\cite{pnp}. The gradient of the data-fidelity term is back-propagated through the decoder $\mathcal{D}(\cdot)$, enforcing consistency directly on the latent variable. 
The update reads
\begin{equation}
\hat{\mathbf{x}} 
= \tilde{\mathbf{z}} - \mathbf{u} 
  - \gamma_t \nabla_{\mathbf{v}=\tilde{\mathbf{z}}-\mathbf{u}}
    \big\|\mathbf{y} - \mathcal{A}(\mathcal{D}(\mathbf{v}))\big\|_2^2,
\label{eq:latent_dc}
\end{equation}
where the gradient involves the Jacobian $J_{\mathcal{D}}(\mathbf{v})$ of the decoder, i.e.,
$\nabla_{\mathbf{v}}\|\mathbf{y}-\mathcal{A}(\mathcal{D}(\mathbf{v}))\|^2 
= J_{\mathcal{D}}(\mathbf{v})^\top \nabla_{\mathcal{D}(\mathbf{v})}\|\mathbf{y}-\mathcal{A}(\cdot)\|^2$. 

After the update in~\eqref{eq:latent_dc}, we perform a subsequent \emph{Re-Encode} step, 
$\mathbf{x} \leftarrow \mathcal{E}(\mathcal{D}(\hat{\mathbf{x}}))$, following the strategy used in DCDP~\cite{dcdp}. This re-encoding step ensures that the updated latent remains on the manifold learned by the LDM, 
since the encoder $\mathcal{E}$ is nonlinear and the model is trained only on latents of the form $\mathbf{z}=\mathcal{E}(\mathbf{x})$ corresponding to valid images.

\paragraph{(2) \textsc{Pixel-DC}: data consistency in pixel space.}
Alternatively, the latent is first decoded to the pixel domain, and the correction is applied directly on the reconstructed image. This is done by minimizing the following objective:

\begin{equation}
\hat{\mathbf{x}} 
\leftarrow \operatorname*{argmin}_{\hat{\mathbf{x}}}
\frac{1}{2\sigma^2}\big\|\mathbf{y}-\mathcal{A}(\hat{\mathbf{x}})\big\|_2^2
+ \frac{\rho}{2}\big\|\hat{\mathbf{x}}-\tilde{\mathbf{x}}+\mathbf{u}\big\|_2^2,
\label{eq:pixel_dc_objective}
\end{equation}
where $\tilde{\mathbf{x}} = \mathcal{D}(\tilde{\mathbf{z}})$. As in \textsc{Latent-DC}, \eqref{eq:pixel_dc_objective} can be solved via a single gradient step:

\begin{equation}
\hat{\mathbf{x}} 
= \tilde{\mathbf{x}} - \mathbf{u} 
  - \gamma_t \nabla_{\mathbf{v}=\tilde{\mathbf{x}}-\mathbf{u}}
    \big\|\mathbf{y} - \mathcal{A}(\mathbf{v})\big\|_2^2.
\label{eq:pixel_dc}
\end{equation}
The corrected image $\hat{\mathbf{x}}$ is then encoded via $\mathbf{x} = \mathcal{E}(\hat{\mathbf{x}})$ for the next diffusion iteration.

Compared to \textsc{Latent-DC}, the \textsc{Pixel-DC} variant is considerably more efficient since it avoids back-propagation through the deep decoder $\mathcal{D}$. In practice, both methods are effective across various inverse problems; however, we find that \textsc{Pixel-DC} yields consistently higher-quality reconstructions. 
Consequently, all quantitative results reported in Table \ref{tab:latent_comparison} correspond to the \textsc{Pixel-DC} variant. We provide the full algorithm in Algorithm~\ref{alg:latent_ddiff}.

\begin{table}[htbp]
  \caption{\textbf{Quantitative evaluation (latent diffusion).} Comparing latent-space methods across 5 linear and 2 nonlinear tasks. We use 100 validation images and report the average metric value. The best and second-best results are distinguished by \textbf{bold} and \underline{underlined} marks, respectively. All tasks are run with a noise of standard deviation $\sigma=0.05$. LatentDDiff achieves competitive performance relative to prior latent methods.}
  \centering
  \captionsetup{font=small}
  \begin{adjustbox}{max width=\columnwidth}
  \scriptsize
  \begin{tabular}{l|l|cccc|cccc}
    \toprule
    & & \multicolumn{4}{c|}{\textbf{FFHQ}} & \multicolumn{4}{c}{\textbf{ImageNet}} \\
    \textbf{Task} & \textbf{Method} & PSNR ($\uparrow$) & SSIM ($\uparrow$) & LPIPS ($\downarrow$) & Residual ($\downarrow$) & PSNR ($\uparrow$) & SSIM ($\uparrow$) & LPIPS ($\downarrow$) & Residual ($\downarrow$) \\
    \midrule

\multirow{4}{*}{\begin{tabular}[c]{@{}l@{}}Super\\Resolution 4×\end{tabular}}
& LatentDDiff & \underline{26.67} & \underline{0.688} & 0.324 & \underline{0.0040} & \textbf{28.65} & 0.664 & 0.411 & \textbf{0.0043} \\
& LatentDAPS & \textbf{27.35} & \textbf{0.800} & \textbf{0.185} & \textbf{0.0038} & 25.01 & \underline{0.669} & \textbf{0.284} & 0.0058 \\
& PSLD       & 24.24 & 0.639 & \underline{0.289} & 0.0060 & \underline{25.40} & \textbf{0.689} & \underline{0.362} & \underline{0.0056} \\
& ReSample   & 23.13 & 0.588 & 0.401 & 0.0065 & 22.57 & 0.567 & 0.382 & 0.0065 \\
\midrule

\multirow{4}{*}{\begin{tabular}[c]{@{}l@{}}Inpainting\\(Box)\end{tabular}}
& LatentDDiff & 20.83 & 0.617 & 0.208 & 0.0192 & \underline{19.10} & 0.551 & \underline{0.337} & \textbf{0.0162} \\
& LatentDAPS & \underline{23.86} & \underline{0.799} & 0.192 & 0.0146 & 17.15 & \underline{0.645} & 0.341 & 0.0372 \\
& PSLD       & \textbf{24.15} & \textbf{0.800} & \textbf{0.163} & \textbf{0.0095} & \textbf{20.08} & \textbf{0.687} & 0.479 & 0.0390 \\
& ReSample   & 19.56 & 0.621 & \underline{0.190} & \underline{0.0131} & 18.13 & 0.625 & \textbf{0.270} & \underline{0.0214} \\
\midrule

\multirow{4}{*}{\begin{tabular}[c]{@{}l@{}}Inpainting\\(Random)\end{tabular}}
& LatentDDiff & 26.81 & 0.643 & 0.239 & \underline{0.0035} & \underline{29.15} & 0.665 & 0.341 & \textbf{0.0038} \\
& LatentDAPS & \textbf{30.52} & \textbf{0.804} & \underline{0.153} & \textbf{0.0033} & 27.31 & \textbf{0.769} & \underline{0.177} & 0.0058 \\
& PSLD       & \underline{30.14} & \underline{0.776} & 0.228 & 0.0037 & \textbf{30.23} & \underline{0.760} & 0.340 & \underline{0.0042} \\
& ReSample   & 29.24 & 0.730 & \textbf{0.145} & 0.0036 & 27.03 & 0.721 & \textbf{0.155} & 0.0095 \\
\midrule

\multirow{4}{*}{\begin{tabular}[c]{@{}l@{}}Gaussian\\Deblurring\end{tabular}}
& LatentDDiff & \underline{27.56} & \underline{0.725} & \underline{0.252} & \underline{0.0027} & \textbf{26.06} & \textbf{0.850} & \underline{0.310} & \textbf{0.0027} \\
& LatentDAPS & \textbf{27.58} & \textbf{0.753} & \textbf{0.248} & \textbf{0.0026} & 24.67 & 0.637 & 0.356 & 0.0067 \\
& PSLD       & 22.98 & 0.600 & 0.331 & 0.0064 & 25.03 & 0.656 & 0.411 & 0.0058 \\
& ReSample   & 25.56 & 0.697 & 0.278 & 0.0045 & \underline{25.45} & \underline{0.683} & \textbf{0.267} & \underline{0.0049} \\
\midrule

\multirow{4}{*}{\begin{tabular}[c]{@{}l@{}}Motion\\Deblurring\end{tabular}}
& LatentDDiff & 26.11 & 0.593 & \underline{0.286} & 0.0028 & 25.54 & 0.637 & 0.330 & 0.0038 \\
& LatentDAPS & \underline{26.88} & \underline{0.799} & 0.299 & \underline{0.0027} & \underline{26.34} & \textbf{0.721} & \underline{0.304} & \underline{0.0034} \\
& PSLD       & 21.81 & 0.642 & 0.357 & 0.0036 & 20.23 & 0.566 & 0.525 & 0.0067 \\
& ReSample   & \textbf{27.21} & \textbf{0.801} & \textbf{0.214} & \textbf{0.0025} & \textbf{26.65} & \underline{0.706} & \textbf{0.244} & \textbf{0.0030} \\
\midrule


\multirow{3}{*}{\begin{tabular}[c]{@{}l@{}}Nonlinear\\Deblurring\end{tabular}}
& LatentDDiff & 27.06 & 0.683 & \underline{0.234} & \textbf{0.0050} & \textbf{26.80} & \textbf{0.711} & \underline{0.265} & \textbf{0.0041} \\
& LatentDAPS & \underline{27.87} & \underline{0.695} & 0.250 & 0.0056 & 25.04 & 0.612 & 0.346 & 0.0057 \\
& ReSample   & \textbf{28.04} & \textbf{0.722} & \textbf{0.199} & \underline{0.0052} & \underline{25.89} & \underline{0.621} & \textbf{0.245} & \underline{0.0044} \\
\midrule

\multirow{3}{*}{\begin{tabular}[c]{@{}l@{}}High\\Dynamic\\Range\end{tabular}}
& LatentDDiff & 22.68 & 0.648 & \underline{0.204} & \textbf{0.0145} & 21.69 & \textbf{0.662} & \textbf{0.201} & \textbf{0.0360} \\
& LatentDAPS & \textbf{25.66} & \textbf{0.737} & 0.255 & \underline{0.0154} & \underline{23.32} & 0.589 & 0.280 & \underline{0.0366} \\
& ReSample   & 25.21 & \underline{0.702} & \textbf{0.197} & 0.0161 & \textbf{24.97} & \underline{0.616} & \underline{0.208} & 0.0381 \\

    \bottomrule
  \end{tabular}
  \end{adjustbox}
  \label{tab:latent_comparison}
\end{table}

\newpage
\section{Statistical Significance}
\label{appendix:stat_sig}
Building on the main results table, we report 95\% confidence intervals for \Method and DAPS in Table~\ref{tab:standard_error}, highlighting the statistical significance of our improvement, compared to DAPS~\cite{daps}. These statistics also serve as empirical evidence for the practical uniqueness of the fixed point established in Theorem~\ref{thm:fp-ddiff}; the narrow confidence intervals across all tasks indicate that independent initializations $\mathbf{x}_T \sim \mathcal{N}(\mathbf{0}, \mathbf{I})$ yield tightly concentrated reconstructions.


\begin{table}[H]
  \caption{\textbf{Quantitative evaluation with confidence intervals.} We show the average metric values over 100 validation images and the corresponding 95\% confidence intervals for \Method and DAPS.
  }
  \vspace{5pt}
  \ra{1.2}
  \centering
  \footnotesize
  \begin{adjustbox}{max width=\textwidth}
  \begin{tabular}{l|l|cccc|cccc}
    \toprule
    & & \multicolumn{4}{c|}{\textbf{FFHQ}} & \multicolumn{4}{c}{\textbf{ImageNet}} \\
    \textbf{Task} & \textbf{Method} & PSNR ($\uparrow$) & SSIM ($\uparrow$) & LPIPS ($\downarrow$) & Residual ($\downarrow$) & PSNR ($\uparrow$) & SSIM ($\uparrow$) & LPIPS ($\downarrow$) & Residual ($\downarrow$) \\
    \midrule
    
    \multirow{2}{*}{\begin{tabular}[c]{@{}l@{}}Super\\Resolution 4×\end{tabular}} 
    & \Method (ours) & \textbf{30.07}{\scriptsize $\pm$0.41} & \textbf{0.824}{\scriptsize $\pm$0.008} & 0.211{\scriptsize $\pm$0.009} & (\textbf{2.85}{\scriptsize $\pm0.06$})$\cdot10^{-3}$ & \textbf{25.81}{\scriptsize $\pm$0.72} & \textbf{0.656}{\scriptsize $\pm$0.029} & 0.396{\scriptsize $\pm$0.029} & (\textbf{3.83}{\scriptsize $\pm0.29$})$\cdot10^{-3}$ \\
    & DAPS & 29.34{\scriptsize$\pm$0.33} & 0.783{\scriptsize$\pm$0.006} & \textbf{0.190}{\scriptsize$\pm$0.006} & (2.97{\scriptsize $\pm0.06$})$\cdot10^{-3}$ & 25.44{\scriptsize $\pm$0.56} & 0.636{\scriptsize $\pm$0.017} & \textbf{0.295}{\scriptsize $\pm$0.011} & (4.77{\scriptsize $\pm0.36$})$\cdot10^{-3}$ \\
    \midrule
    
    \multirow{2}{*}{\begin{tabular}[c]{@{}l@{}}Inpainting\\(Box)\end{tabular}} 
    & \Method (ours) & \textbf{24.88}{\scriptsize $\pm$0.50} & \textbf{0.831}{\scriptsize $\pm$0.005} & \textbf{0.110}{\scriptsize $\pm$0.005} & (\textbf{7.75}{\scriptsize $\pm9.21$})$\cdot10^{-3}$ & 21.15{\scriptsize $\pm$0.68} & \textbf{0.743}{\scriptsize $\pm$0.009} & 0.240{\scriptsize $\pm$0.011} & (\textbf{1.19}{\scriptsize $\pm0.29$})$\cdot10^{-2}$ \\
    & DAPS & 24.52{\scriptsize $\pm$0.40} & 0.742{\scriptsize $\pm$0.006} & 0.174{\scriptsize $\pm$0.006} & (9.86{\scriptsize $\pm6.76$})$\cdot10^{-3}$ & \textbf{21.22}{\scriptsize $\pm$0.70} & 0.714{\scriptsize $\pm$0.007} & \textbf{0.230}{\scriptsize $\pm$0.011} & (1.50{\scriptsize $\pm0.24$})$\cdot10^{-2}$ \\
    \midrule
    
    \multirow{2}{*}{\begin{tabular}[c]{@{}l@{}}Inpainting\\(Random)\end{tabular}} 
    & \Method (ours) & \textbf{33.08}{\scriptsize $\pm$0.37} & \textbf{0.877}{\scriptsize $\pm$0.004} & \textbf{0.050}{\scriptsize $\pm$0.003} & (\textbf{2.05}{\scriptsize $\pm0.15$})$\cdot10^{-2}$ & \textbf{28.39}{\scriptsize $\pm$0.76} & \textbf{0.758}{\scriptsize $\pm$0.019} & \textbf{0.136}{\scriptsize $\pm$0.018} & (\textbf{2.41}{\scriptsize $\pm0.35$})$\cdot10^{-2}$ \\
    & DAPS & 30.76{\scriptsize $\pm$0.27} & 0.801{\scriptsize $\pm$0.005} & 0.156{\scriptsize $\pm$0.003} & (2.93{\scriptsize $\pm0.16$})$\cdot10^{-2}$ & 27.32{\scriptsize $\pm$0.63} & 0.725{\scriptsize $\pm$0.013} & 0.189{\scriptsize $\pm$0.010} & (7.88{\scriptsize $\pm0.70$})$\cdot10^{-2}$ \\
    \midrule
    
    \multirow{2}{*}{\begin{tabular}[c]{@{}l@{}}Gaussian\\Deblurring\end{tabular}} 
    & \Method (ours) & 28.87{\scriptsize $\pm$0.43} & \textbf{0.800}{\scriptsize $\pm$0.010} & \textbf{0.119}{\scriptsize $\pm$0.005} & (\textbf{2.60}{\scriptsize $\pm0.07$})$\cdot10^{-3}$ & 22.29{\scriptsize $\pm$0.84} & 0.471{\scriptsize $\pm$0.039} & 0.415{\scriptsize $\pm$0.035} & (\textbf{4.68}{\scriptsize $\pm0.39$})$\cdot10^{-3}$ \\
    & DAPS & \textbf{29.63}{\scriptsize $\pm$0.36} & 0.789{\scriptsize $\pm$0.006} & 0.177{\scriptsize $\pm$0.005} & (2.67{\scriptsize $\pm0.06$})$\cdot10^{-3}$ & \textbf{25.90}{\scriptsize $\pm$0.64} & \textbf{0.658}{\scriptsize $\pm$0.020} & \textbf{0.269}{\scriptsize $\pm$0.011} & (8.46{\scriptsize $\pm0.31$})$\cdot10^{-3}$ \\
    \midrule
    
    \multirow{2}{*}{\begin{tabular}[c]{@{}l@{}}Motion\\Deblurring\end{tabular}} 
    & \Method (ours) & 28.24{\scriptsize $\pm$0.38} & 0.785{\scriptsize $\pm$0.009} & \textbf{0.129}{\scriptsize $\pm$0.005} & (\textbf{5.82}{\scriptsize $\pm0.27$})$\cdot10^{-3}$ & 24.16{\scriptsize $\pm$0.65} & 0.585{\scriptsize $\pm$0.027} & 0.242{\scriptsize $\pm$0.014} & (\textbf{7.97}{\scriptsize $\pm0.60$})$\cdot10^{-3}$ \\
    & DAPS & \textbf{29.17}{\scriptsize $\pm$0.36} & \textbf{0.797}{\scriptsize $\pm$0.007} & 0.186{\scriptsize $\pm$0.005} & (5.91{\scriptsize $\pm0.28$})$\cdot10^{-3}$ & \textbf{26.61}{\scriptsize $\pm$0.65} & \textbf{0.710}{\scriptsize $\pm$0.020} & \textbf{0.241}{\scriptsize $\pm$0.011} & (8.58{\scriptsize $\pm0.25$})$\cdot10^{-3}$ \\
    \midrule
    
    \multirow{2}{*}{\begin{tabular}[c]{@{}l@{}}Phase \\Retrieval\end{tabular}} 
    & \Method (ours) & \textbf{29.94}{\scriptsize $\pm$0.88} & \textbf{0.816}{\scriptsize $\pm$0.025} & \textbf{0.120}{\scriptsize $\pm$0.019} & (\textbf{4.02}{\scriptsize $\pm0.22$})$\cdot10^{-3}$ & 18.54{\scriptsize $\pm$1.23} & \textbf{0.494}{\scriptsize $\pm$0.045} & \textbf{0.262}{\scriptsize $\pm$0.029} & (\textbf{5.92}{\scriptsize $\pm0.24$})$\cdot10^{-3}$ \\
    & DAPS & 29.60{\scriptsize $\pm$0.88} & 0.768{\scriptsize $\pm$0.020} & 0.182{\scriptsize $\pm$0.018} & (4.94{\scriptsize $\pm0.25$})$\cdot10^{-3}$ & \textbf{20.23}{\scriptsize $\pm$1.27} & 0.449{\scriptsize $\pm$0.044} & 0.397{\scriptsize $\pm$0.032} & (8.53{\scriptsize $\pm0.25$})$\cdot10^{-3}$ \\
    \midrule
    
    \multirow{2}{*}{\begin{tabular}[c]{@{}l@{}}Nonlinear\\Deblurring\end{tabular}} 
    & \Method (ours) & \textbf{31.48}{\scriptsize $\pm$0.29} & \textbf{0.873}{\scriptsize $\pm$0.005} & \textbf{0.120}{\scriptsize $\pm$0.006} & (\textbf{2.74}{\scriptsize $\pm0.07$})$\cdot10^{-3}$ & \textbf{29.68}{\scriptsize $\pm$0.62} & \textbf{0.805}{\scriptsize $\pm$0.016} & \textbf{0.207}{\scriptsize $\pm$0.019} & (\textbf{3.55}{\scriptsize $\pm0.18$})$\cdot10^{-3}$ \\
    & DAPS & 28.45{\scriptsize $\pm$0.38} & 0.764{\scriptsize $\pm$0.007} & 0.188{\scriptsize $\pm$0.006} & (4.16{\scriptsize $\pm0.43$})$\cdot10^{-3}$ & 27.28{\scriptsize $\pm$0.62} & 0.718{\scriptsize $\pm$0.017} & 0.213{\scriptsize $\pm$0.011} & (4.87{\scriptsize $\pm0.33$})$\cdot10^{-3}$ \\
    \midrule
    
    \multirow{2}{*}{\begin{tabular}[c]{@{}l@{}}HDR\end{tabular}} 
    & \Method (ours) & 26.05{\scriptsize $\pm$0.68} & \textbf{0.873}{\scriptsize $\pm$0.007} & \textbf{0.129}{\scriptsize $\pm$0.009} & (\textbf{4.59}{\scriptsize $\pm$0.27})$\cdot10^{-2}$ & \textbf{26.50}{\scriptsize $\pm$0.56} & 0.800{\scriptsize $\pm$0.018} & \textbf{0.108}{\scriptsize $\pm$0.014} & (\textbf{5.41}{\scriptsize $\pm$0.27})$\cdot10^{-2}$ \\
    & DAPS & \textbf{27.39}{\scriptsize $\pm$0.56} & 0.846{\scriptsize $\pm$0.010} & 0.163{\scriptsize $\pm$0.012} & (5.05{\scriptsize $\pm0.18$})$\cdot10^{-2}$ & 26.10{\scriptsize $\pm$0.79} & \textbf{0.825}{\scriptsize $\pm$0.021} & 0.171{\scriptsize $\pm$0.019} & (7.17{\scriptsize $\pm0.75$})$\cdot10^{-2}$ \\
    \bottomrule

  \end{tabular}
  \end{adjustbox}
  \label{tab:standard_error}
\end{table}

\section{Relating \Method and DiffPIR}
\label{appendix:relation}
Here we show the equivalence of \Method and DiffPIR (see Algorithm \ref{alg:diffpir}) under specific parameter choices. DiffPIR introduces $\zeta$ as a hyperparameter controlling the stochasticity of the algorithm and $\bar{\sigma}_t$ as a parameter controlling the strength of the likelihood step. Our algorithm is equivalent to the linearized version of DiffPIR when:


\begin{enumerate}[label=\roman*.]
    \item $\sigma_t = \sqrt{\zeta \cdot (1-\bar{\alpha}_{t-1})}$; 
    \item $\gamma_t=\frac{\bar{\sigma}_t^2}{2\lambda\sigma^2}$ where $\sigma$ is the measurement noise standard deviation;
    \item We remove early the relaxation parameter $t_0$;
    \item We remove the dual updates.
\end{enumerate}

\begin{algorithm}[H]
\caption{DiffPIR}
\label{alg:diffpir}
\begin{algorithmic}[1]
\REQUIRE $T$, $\mathcal{A}(\cdot)$, $\{ \bar{\sigma}_t \}_{t=1}^{T}$, $\{ \bar{\alpha}_t \}_{t=1}^{T}$, $\mathbf{s}_\theta$,  $\mathbf{y}$, $\zeta$, $\lambda$, $\sigma$

\STATE Initialize $\mathbf{x}_T \sim \mathcal{N}(\mathbf{0}, \mathbf{I})$.
\FOR{$t = T-1$ $\textbf{to}$ $0$}
    \STATE $\mathbf{x}_0^{(t)} \leftarrow \frac{1}{\sqrt{\bar{\alpha}_t}} \left(\mathbf{x}_t + (1-\bar{\alpha}_t) \mathbf{s}_\theta(\mathbf{x}_t, t)\right)$ 
    \STATE $\hat{\mathbf{x}}_0^{(t)} \leftarrow  \mathbf{x}_0^{(t)} - \frac{\bar{\sigma}_t^2}{2\lambda\sigma^2} \nabla_{\mathbf{x}_0^{(t)}} \|\mathbf{y} - \mathcal{A}( \mathbf{x}_0^{(t)} )\|^2$ 
    \STATE $\hat{\epsilon}\leftarrow \frac{1}{\sqrt{1-\bar{\alpha}_t}} \left( \mathbf{x}_t - \sqrt{\bar{\alpha}_t}\hat{\mathbf{x}}_0^{(t)}  \right)$
    \STATE $\epsilon \sim \mathcal{N}(\mathbf{0}, \mathbf{I})$
    \STATE $\mathbf{x}_{t-1} \leftarrow\sqrt{\bar{\alpha}_{t-1}} \hat{\mathbf{x}}_0^{(t)} + \sqrt{1-\bar{\alpha}_{t-1}}(\sqrt{1-\zeta}\hat{\epsilon}+\sqrt{\zeta}\epsilon_t)$ 
\ENDFOR

\RETURN $\mathbf{x}_0$
\end{algorithmic}
\end{algorithm}

\newpage

\section{\Method Variants Algorithms}
\label{appendix:variants}

This section presents the precise algorithms corresponding to the three variants of \Method discussed in the ablation study. Note that DDiff-HQS \ref{alg:ddiff_hqs} is equivalent to DiffPIR \ref{alg:diffpir} if we satisfy (i.-iii.) from Appendix~\ref{appendix:relation}.

\begin{algorithm}[H]
\caption{Diff-PnP-HQS (no noise, no $\mathbf{u}$)}
\label{alg:diff-pnp-hqs}
\begin{algorithmic}[1]
\REQUIRE $T$, $\mathcal{A}(\cdot)$, $\{ \sigma_t \}_{t=1}^{T}$, $\{ \bar{\alpha}_t \}_{t=1}^{T}$, $\mathbf{s}_\theta$,  $\mathbf{y}$, $\{\gamma_t\}_{t=1}^T$, $t_0$

\STATE Initialize $\mathbf{x}_T \sim \mathcal{N}(\mathbf{0}, \mathbf{I})$.
\FOR{$t = T-1$ $\textbf{to}$ $0$}
    \STATE $\mathbf{z} \leftarrow \frac{1}{\sqrt{\bar{\alpha}_t}} \left(\mathbf{x}_t + (1-\bar{\alpha}_t) \mathbf{s}_\theta(\mathbf{x}_t, t)\right)$ 
    \STATE $\mathbf{x}_{t-1} \leftarrow  \mathbf{z} - \gamma_t \nabla_{\mathbf{z}} \|\mathbf{y} - \mathcal{A}( \mathbf{z} )\|^2$ 
\ENDFOR

\RETURN $\mathbf{x}_0$
\end{algorithmic}
\end{algorithm}

\begin{algorithm}[H]
\caption{Diff-PnP-ADMM (no noise, with $\mathbf{u}$)}
\label{alg:diff-pnp-admm}
\begin{algorithmic}[1]
\REQUIRE $T$, $\mathcal{A}(\cdot)$, $\{ \sigma_t \}_{t=1}^{T}$, $\{ \bar{\alpha}_t \}_{t=1}^{T}$, $\mathbf{s}_\theta$,  $\mathbf{y}$, $\{\gamma_t\}_{t=1}^T$, $t_0$

\STATE Initialize $\mathbf{x}_T \sim \mathcal{N}(\mathbf{0}, \mathbf{I}), \mathbf{u} = \mathbf{0} $.
\FOR{$t = T-1$ $\textbf{to}$ $0$}
    \STATE $\mathbf{z} \leftarrow \frac{1}{\sqrt{\bar{\alpha}_t}} \left(\mathbf{x}_t+\mathbf{u} + (1-\bar{\alpha}_t) \mathbf{s}_\theta(\mathbf{x}_t+\mathbf{u}, t)\right)$ 
    \STATE $\mathbf{x}_{t-1} \leftarrow  \mathbf{z}-\mathbf{u} - \gamma_t \nabla_{\mathbf{v}=\mathbf{z}-\mathbf{u}} \|\mathbf{y} - \mathcal{A}( \mathbf{v} )\|^2$ 
    \STATE $\mathbf{u} \leftarrow \mathbf{u} + \mathbf{x}_{t-1} - \mathbf{z}$
\ENDFOR

\RETURN $\mathbf{x}_0$
\end{algorithmic}
\end{algorithm}

\begin{algorithm}[H]
\caption{\Method-HQS (with noise, no $\mathbf{u}$)}
\label{alg:ddiff_hqs}
\begin{algorithmic}[1]
\REQUIRE $T$, $\mathcal{A}(\cdot)$, $\{ \sigma_t \}_{t=1}^{T}$, $\{ \bar{\alpha}_t \}_{t=1}^{T}$, $\mathbf{s}_\theta$,  $\mathbf{y}$, $\{\gamma_t\}_{t=1}^T$, $t_0$

\STATE Initialize $\mathbf{x}_T \sim \mathcal{N}(\mathbf{0}, \mathbf{I})$.
\FOR{$t = T-1$ $\textbf{to}$ $0$}
    \STATE $\mathbf{z} \leftarrow \frac{1}{\sqrt{\bar{\alpha}_t}} \left(\mathbf{x}_t + (1-\bar{\alpha}_t) \mathbf{s}_\theta(\mathbf{x}_t, t)\right)$ 
    \STATE $\mathbf{x} \leftarrow  \mathbf{z} - \gamma_t \nabla_{\mathbf{z}} \|\mathbf{y} - \mathcal{A}( \mathbf{z} )\|^2$ 
    \STATE $\hat{\epsilon}\leftarrow \frac{1}{\sqrt{1-\bar{\alpha}_t}} \left( \mathbf{x}_t - \sqrt{\bar{\alpha}_t}\cdot\mathbf{x}  \right)$
    \STATE $\epsilon \sim \mathcal{N}(\mathbf{0}, \mathbf{I}) \:\textbf{if} \:t>t_0\:\textbf{else} \: \epsilon=0$
    \STATE $\mathbf{x}_{t-1} \leftarrow\sqrt{\bar{\alpha}_{t-1}} \cdot\mathbf{x} + \sqrt{1-\bar{\alpha}_{t-1}-\sigma_t^2} \cdot \hat{\epsilon} + \sigma_t \epsilon$ 
\ENDFOR

\RETURN $\mathbf{x}_0$
\end{algorithmic}
\end{algorithm}

\section{Inverse Problems Setup}
\label{sec:inverse_problem_setup}
Most inverse problems are implemented using the same approach described in \cite{dps}, except for the HDR task which follows the setup in \cite{daps}. We set a fixed random seed for inpainting, motion deblurring, and nonlinear deblurring for fair comparison. Specific parameters are defined as follows:
\begin{itemize}
    \item Super resolution: $4\times$ downsampling factor
    \item Inpainting: $128\times128$ box mask and $70\%$ random mask
    \item Gaussian \& motion deblurring: $61\times61$ kernel size with standard deviations of $3.0$ and $0.5$, respectively
    \item Phase retrieval: oversampling with ratio $k/n$ where $k=2$ and $n=8$
    \item Nonlinear deblurring: blur kernel generated using \cite{nonlinear-deblur}
    \item HDR: $2\times$ dynamic range
\end{itemize}

\newpage
\section{Hyperparameter Choices}
\label{supp:hyperparam}

The step size $\gamma_t$ for the measurement step is defined by a step function below:
\begin{center}
$\gamma_t = \gamma_0 \cdot f(t_\gamma)$ where $f(t_\gamma) = 
\begin{cases} 
a & \text{for } t > t_\gamma, \\
b & \text{for } t \leq t_\gamma.
\end{cases}
$
\end{center}
To enhance sample quality, the parameter $\gamma_t$ is reduced at time step $t_\gamma$. This adjustment encourages the traversal to relax toward the prior manifold while diverging from the likelihood manifold. As a result, the generated samples exhibit reduced noise and improved visual fidelity. For all pixel-space diffusion tasks, $a=3.3$ and $b=0.1$ worked well empirically. For latent-space diffusion tasks, we report the exact values in Table \ref{tab:parameters_ffhq} and \ref{tab:parameters_imagenet}.

We used two different choices of $\sigma_t$ in the main experiments. 

\begin{itemize}
    \item $\sigma_t=\sqrt{(1-\bar{\alpha}_{t-1})/(1-\bar{\alpha}_t)}\sqrt{1-\bar{\alpha}_t/\bar{\alpha}_{t-1}}$ 
    (this choice allows the generative process to become a DDPM) 
    was used for Gaussian deblurring, motion deblurring, nonlinear deblurring, and HDR. 
    \item $\sigma_t=\sqrt{1-\bar{\alpha}_{t-1}}$ was used for phase retrieval, inpainting (box and random), and super resolution. 
\end{itemize}

The time threshold $t_0$ controls the reverse diffusion step to switch from stochastic to deterministic. 

We report the hyperparameters used in the main experiments in Table~\ref{tab:parameters_ffhq} and \ref{tab:parameters_imagenet}.

\begin{table}[H]
\centering
\small
\caption{\textbf{\Method hyperparameter settings} for different tasks on FFHQ.}
\label{tab:parameters_ffhq}
\resizebox{\textwidth}{!}{
\begin{tabular}{l|l|cccccccc}
\toprule
\textbf{Algorithm} & \textbf{Task} & \textbf{Super Res. $4\times$} & \textbf{Inpaint (Box)} & \textbf{Inpaint (Rand.)} & \textbf{Gaussian Deblur} & \textbf{Motion Deblur} & \textbf{Phase Retrieval} & \textbf{Nonlinear Deblur} & \textbf{HDR} \\
\midrule
\multirow{3}{*}{DDiff} 
& $\gamma_0$ & 18 & 30 & 50 & 2.9 & 2.9 & 38 & 2.5 & 3.5 \\
& $t_\gamma$ & 90 & 90 & 90 & 90 & 90 & 90 & 90 & 90 \\
& $t_0$ & 1 & 1 & 1 & 50 & 80 & 1 & 120 & 120 \\
\midrule
\multirow{5}{*}{LatentDDiff} 
& $\gamma_0$ & 100 & 200 & 400 & 60 & 60 & - & 500 & 150 \\
& $t_\gamma$ & 90 & 600 & 200 & 90 & 150 & - & 90 & 600 \\
& $t_0$ & 500 & 1 & 800 & 200 & 200 & - & 200 & 1 \\
& $a$ & 3.3 & 25 & 30 & 3.3 & 3.3 & - & 3.3 & 30 \\
& $b$ & 0.1 & 0.15 & 0.1 & 0.1 & 0.1 & - & 0.1 & 0.15 \\
\bottomrule
\end{tabular}
}
\end{table}

\begin{table}[H]
\centering
\small
\caption{\textbf{\Method hyperparameter settings} for different tasks on ImageNet.}
\label{tab:parameters_imagenet}
\resizebox{\textwidth}{!}{
\begin{tabular}{l|l|cccccccc}
\toprule
\textbf{Algorithm} & \textbf{Task} & \textbf{Super Res. $4\times$} & \textbf{Inpaint (Box)} & \textbf{Inpaint (Rand.)} & \textbf{Gaussian Deblur} & \textbf{Motion Deblur} & \textbf{Phase Retrieval} & \textbf{Nonlinear Deblur} & \textbf{HDR} \\
\midrule
\multirow{3}{*}{DDiff} 
& $\gamma_0$ & 18 & 50 & 50 & 1.8 & 1.5 & 38 & 2.5 & 3.8 \\
& $t_\gamma$ & 90 & 500 & 90 & 90 & 90 & 90 & 90 & 90 \\
& $t_0$ & 1 & 1 & 1 & 50 & 80 & 1 & 120 & 100 \\
\midrule
\multirow{5}{*}{LatentDDiff} 
& $\gamma_0$ & 150 & 400 & 400 & 150 & 100 & - & 100 & 250 \\
& $t_\gamma$ & 600 & 600 & 600 & 600 & 90 & - & 90 & 600 \\
& $t_0$ & 1 & 1 & 1 & 1 & 200 & - & 200 & 200 \\
& $a$ & 50 & 30 & 50 & 50 & 3.3 & - & 3.3 & 25 \\
& $b$ & 0.15 & 0.1 & 0.15 & 0.15 & 0.1 & - & 0.1 & 0.15 \\
\bottomrule
\end{tabular}
}
\end{table}

\section{Baseline Details}

We follow the baseline configurations reported in DAPS~\cite{daps} for all methods we compare against: DPS~\cite{dps}, DDRM~\cite{ddrm}, DCDP~\cite{dcdp}, DiffPIR~\cite{diffpir}, RED-diff~\cite{red-diff}, PSLD~\cite{psld}, and ReSample~\cite{resample}. For DAPS and DMPlug~\cite{dmplug}, we use the authors' official implementations with default hyperparameters. All baselines use the same pretrained models (as specified in Sec. \ref{sec:experimental_setup}) and forward operators (Sec. \ref{sec:inverse_problem_setup}).

In our experiments, we use DDiff-1k for all inverse problems. In contrast, DAPS uses DAPS-1k for linear tasks and DAPS-4k for nonlinear tasks by default, and we retain these defaults (including hyperparameters) in our comparisons. Under these settings, DDiff attains higher reconstruction quality (PSNR/LPIPS; see Table \ref{tab:comparison}) while requiring roughly half the sampling time on linear tasks and about one-sixth the sampling time on nonlinear tasks (see Fig. \ref{fig:sample_time}).

\section{More Qualitative Results}

\begin{figure}[htbp]
    \centering
    \includegraphics[height=0.98\textheight, keepaspectratio]{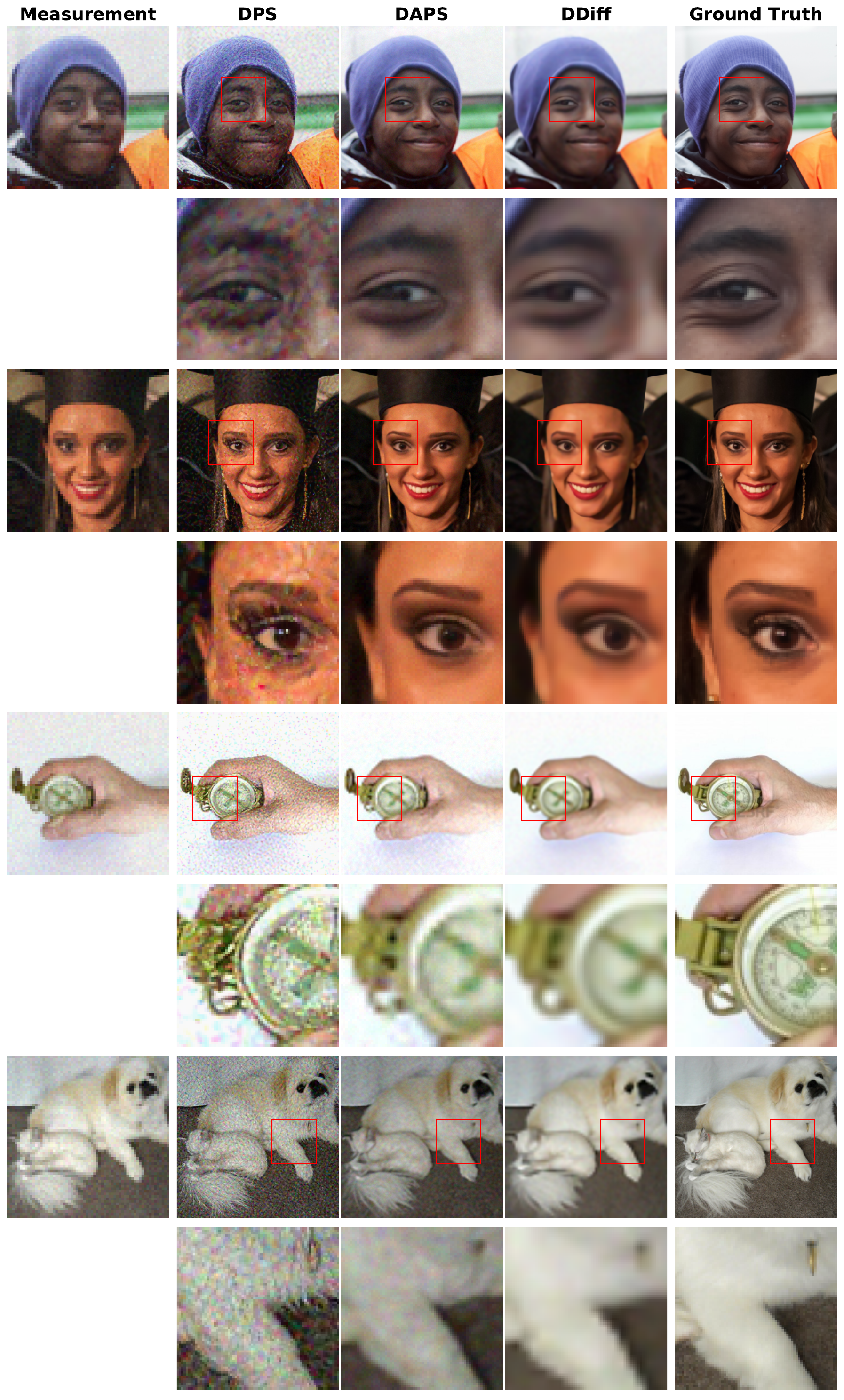}
    \caption{\textbf{Visual comparison of \Method and baselines} on super resolution task.}
    \label{fig:superres_comparison}
\end{figure}

\begin{figure}[htbp]
    \centering
    \includegraphics[height=0.98\textheight, keepaspectratio]{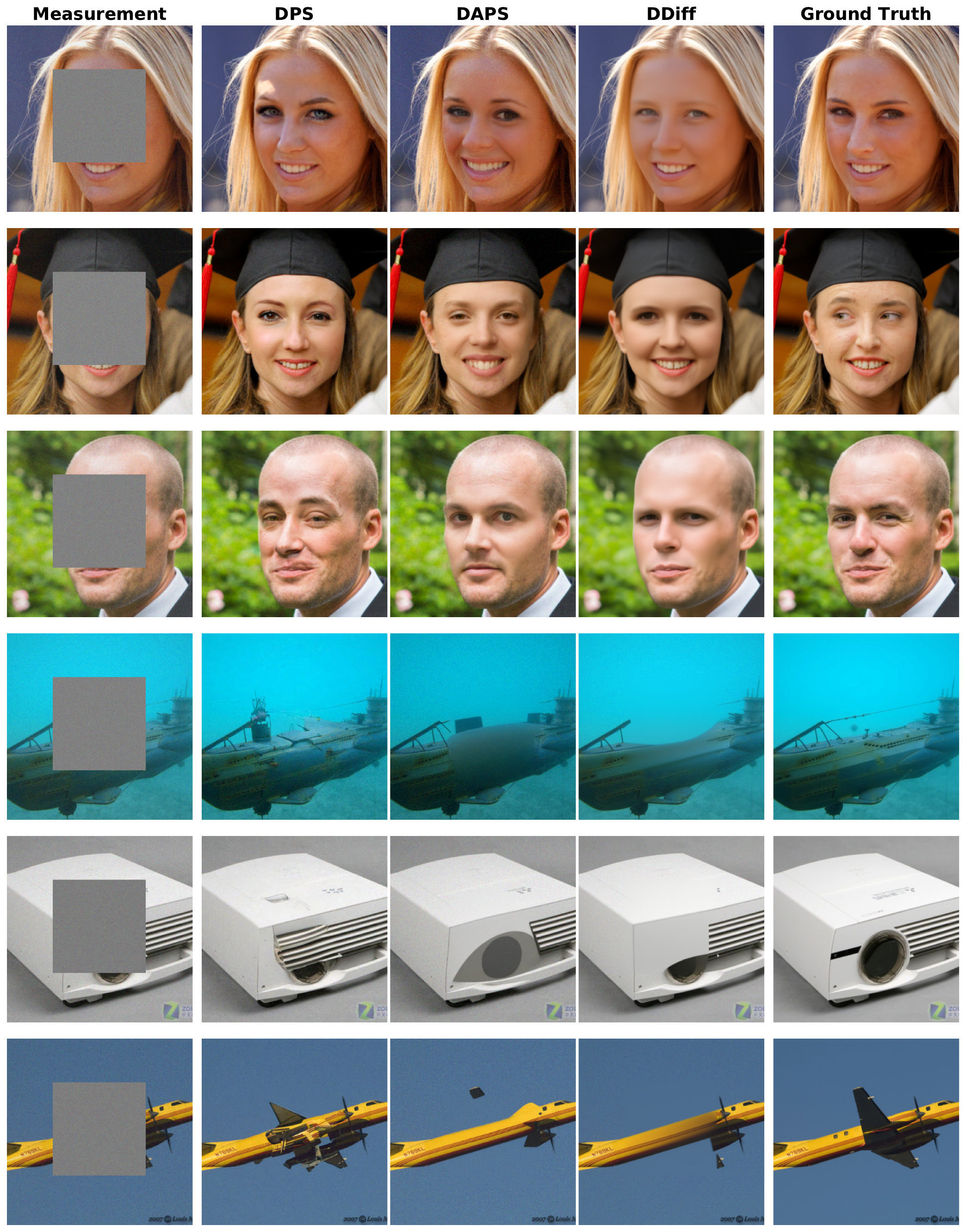}
    \caption{\textbf{Visual comparison of \Method and baselines} on box inpainting task.}
    \label{fig:boxinpaint_comparison}
\end{figure}

\begin{figure}[htbp]
    \centering
    \includegraphics[height=0.98\textheight, keepaspectratio]{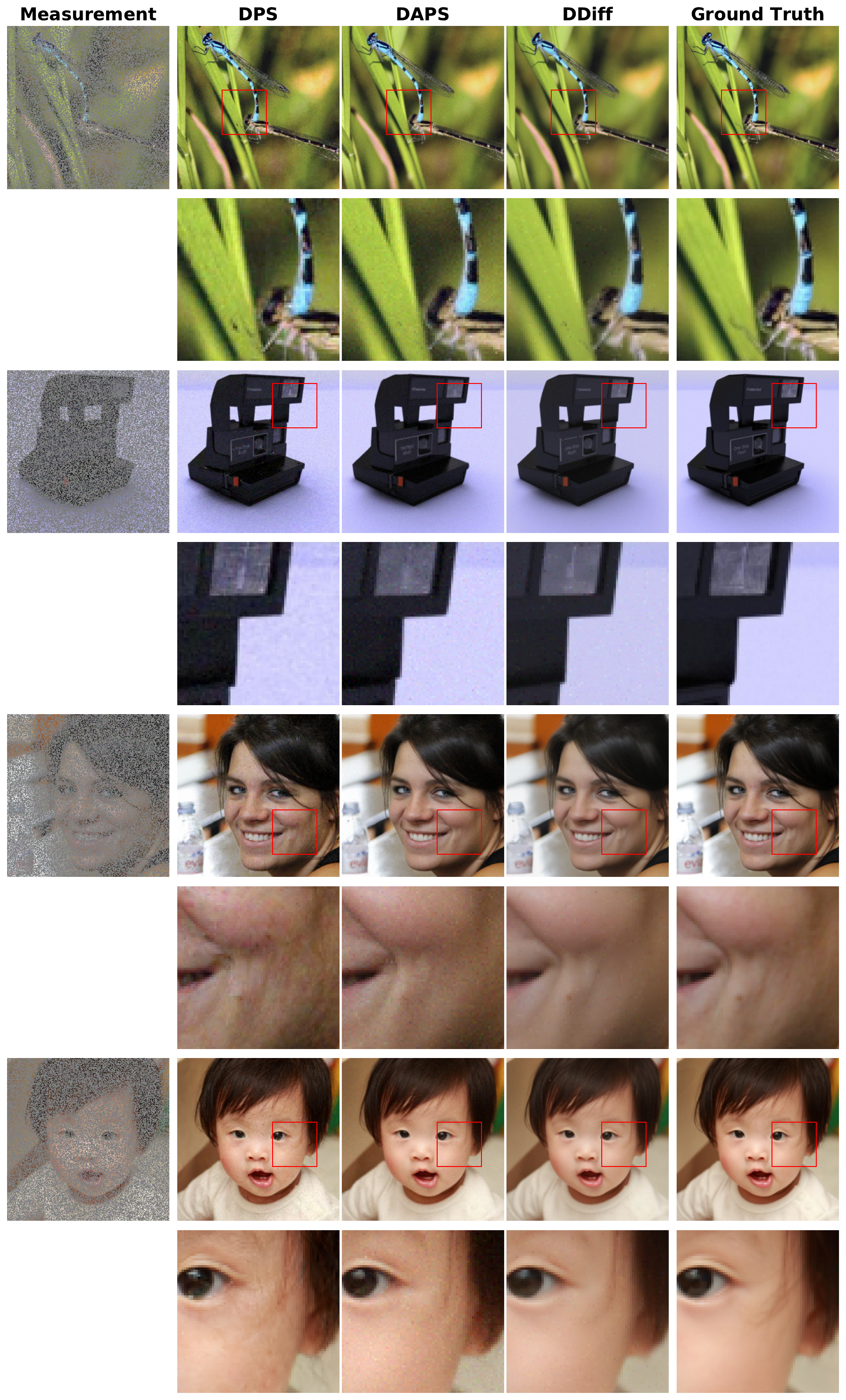}
    \caption{\textbf{Visual comparison of \Method and baselines} on random inpainting task.}
    \label{fig:randinpaint_comparison}
\end{figure}

\begin{figure}[htbp]
    \centering
    \includegraphics[height=0.98\textheight, keepaspectratio]{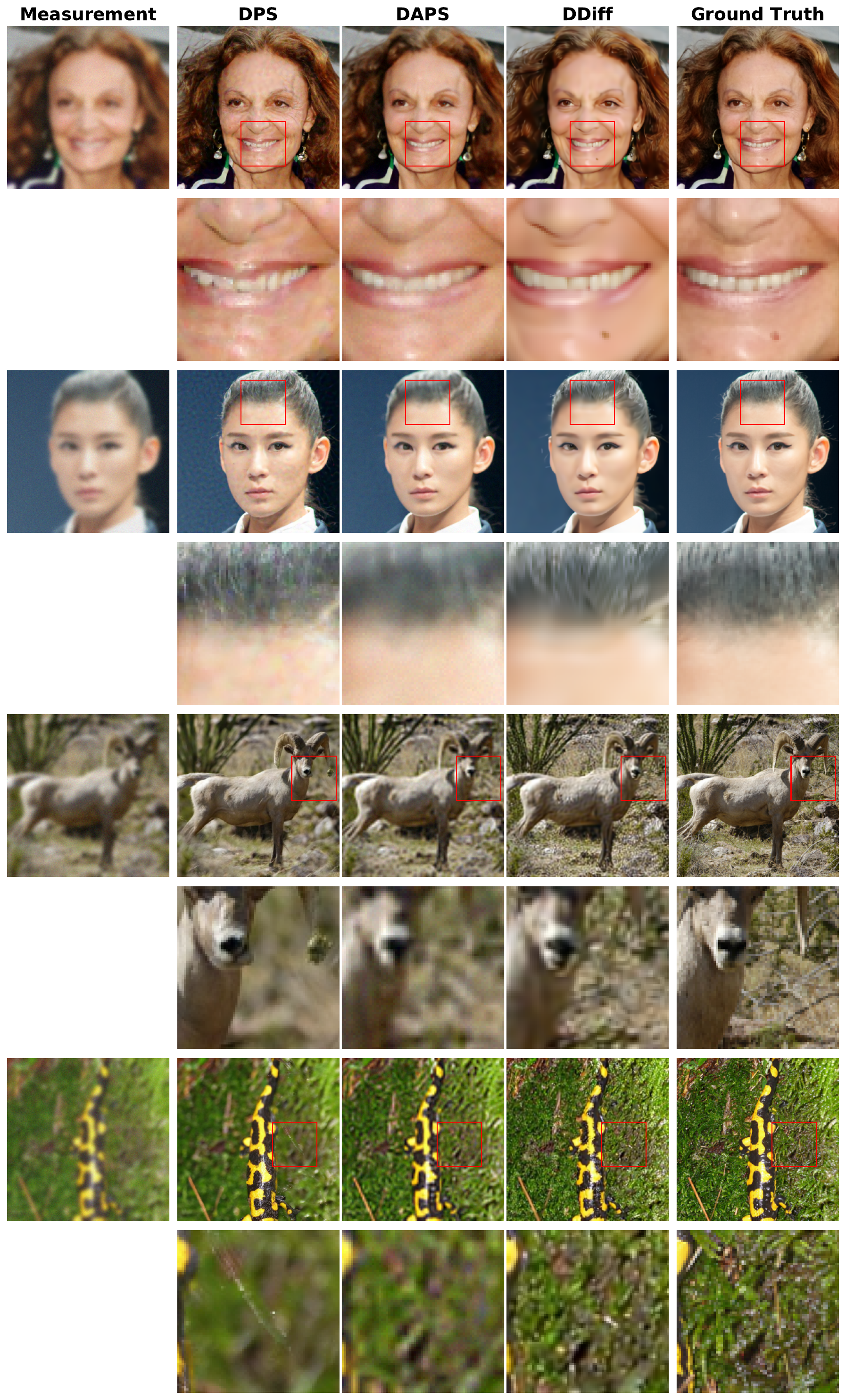}
    \caption{\textbf{Visual comparison of \Method and baselines} on Gaussian deblurring task.}
    \label{fig:deconv_comparison}
\end{figure}

\begin{figure}[htbp]
    \centering
    \includegraphics[height=0.98\textheight, keepaspectratio]{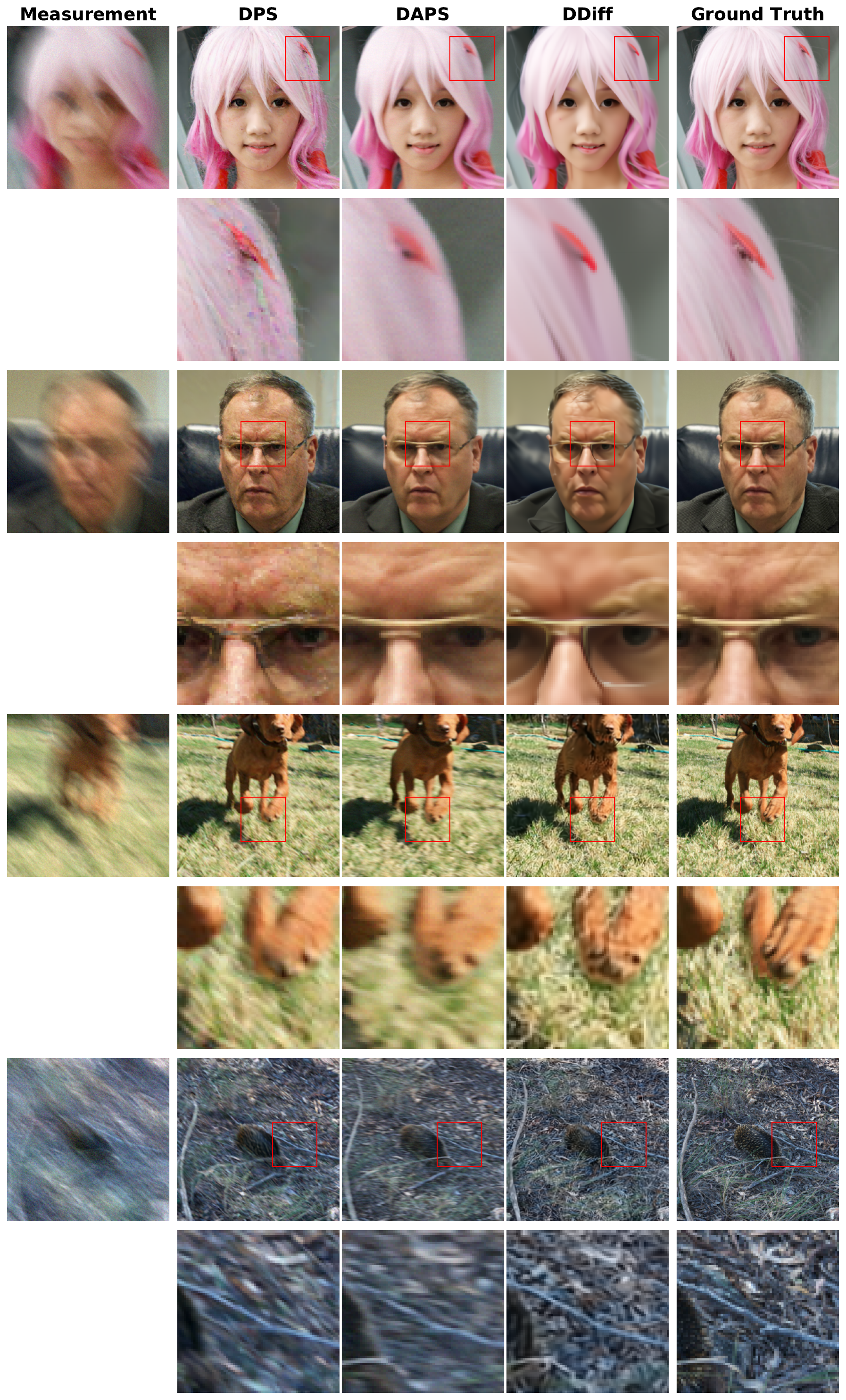}
    \caption{\textbf{Visual comparison of \Method and baselines} on motion deblurring task.}
    \label{fig:motion_deblur_comparison}
\end{figure}

\begin{figure}[htbp]
    \centering
    \includegraphics[height=0.98\textheight, keepaspectratio]{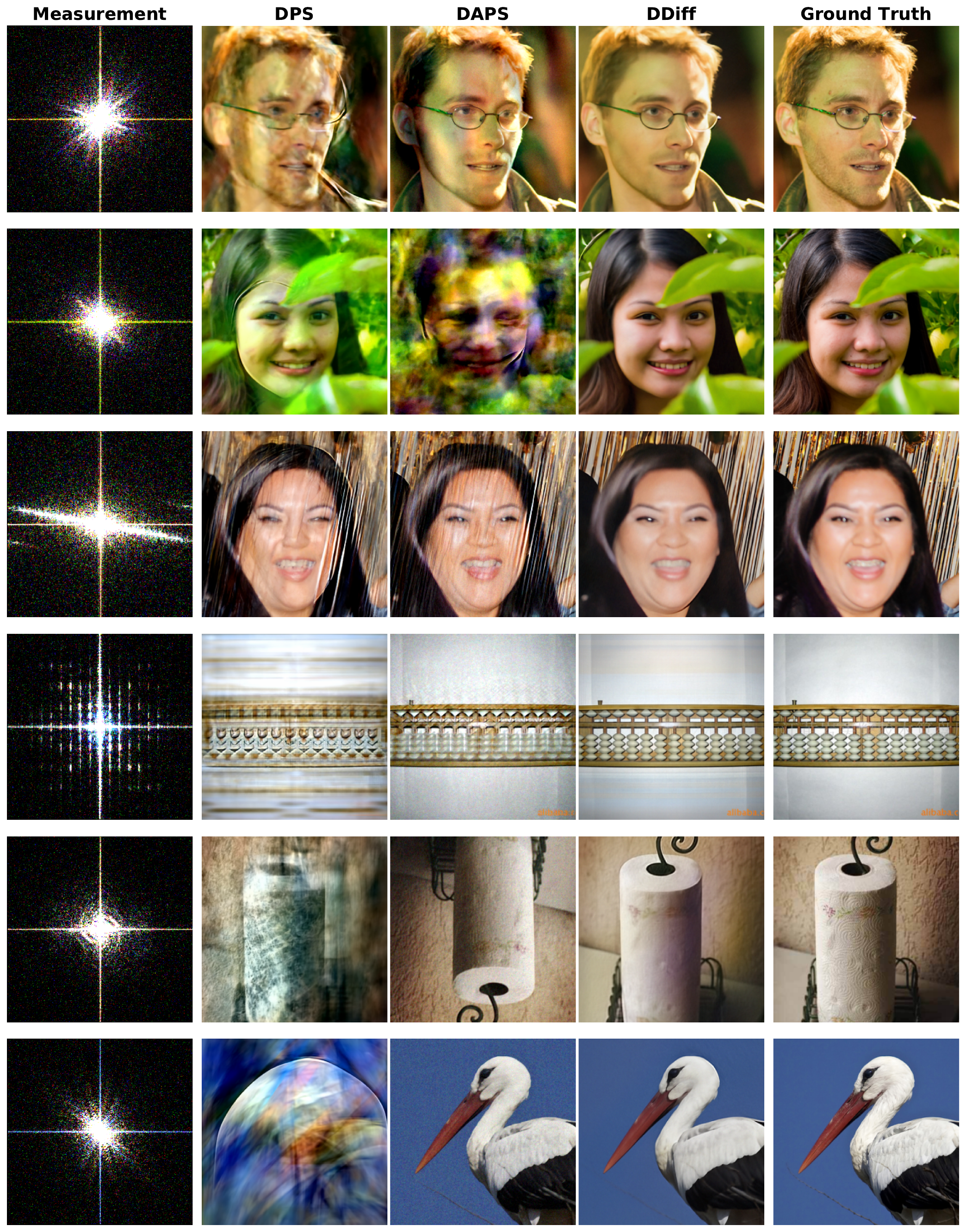}
    \caption{\textbf{Visual comparison of \Method and baselines} on phase retrieval task.}
    \label{fig:pr_comparison}
\end{figure}

\begin{figure}[htbp]
    \centering
    \includegraphics[height=0.98\textheight, keepaspectratio]{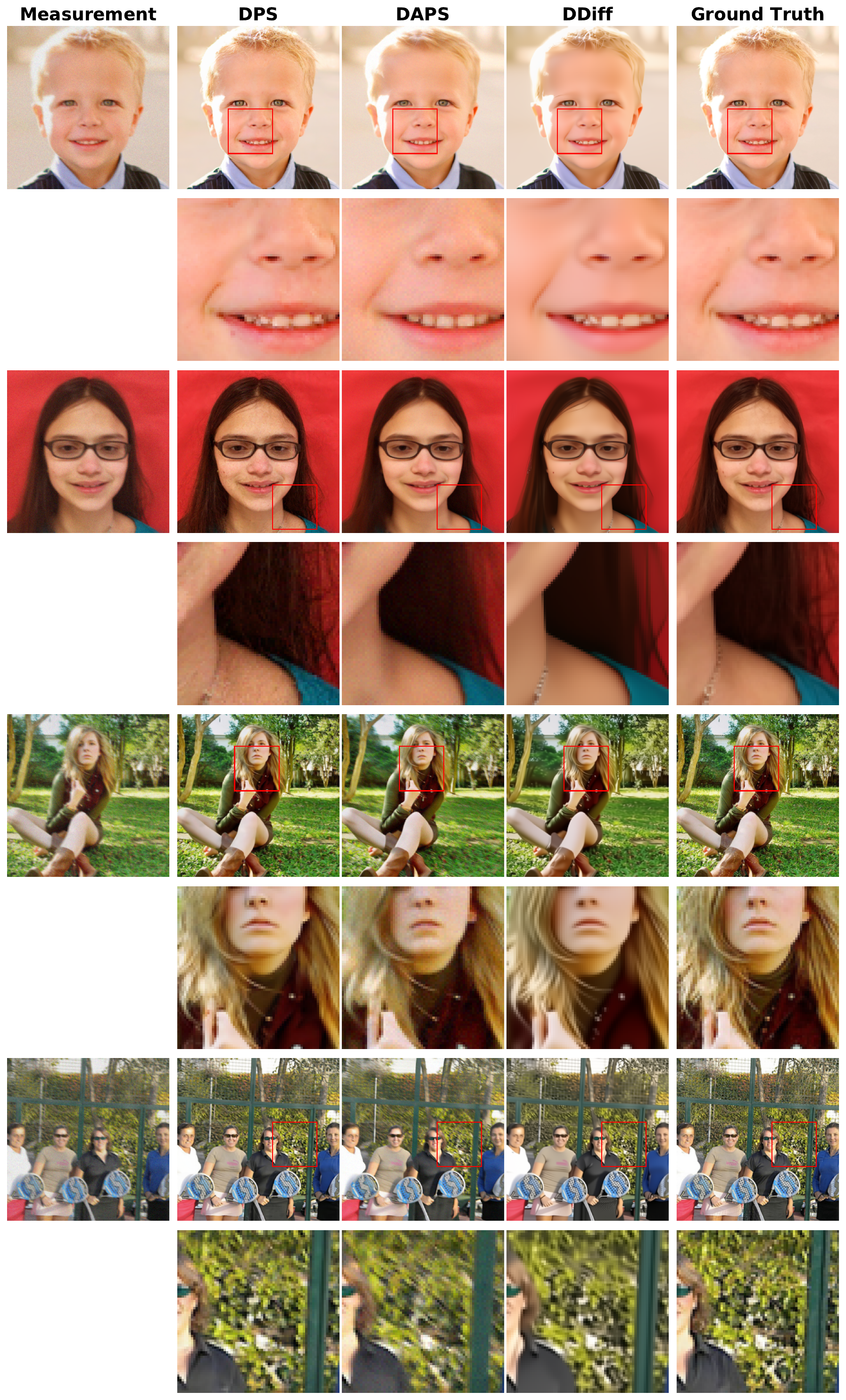}
    \caption{\textbf{Visual comparison of \Method and baselines} on nonlinear deblurring task.}
    \label{fig:nonlinear_blur_comparison}
\end{figure}

\begin{figure}[htbp]
    \centering
    \includegraphics[height=0.98\textheight, keepaspectratio]{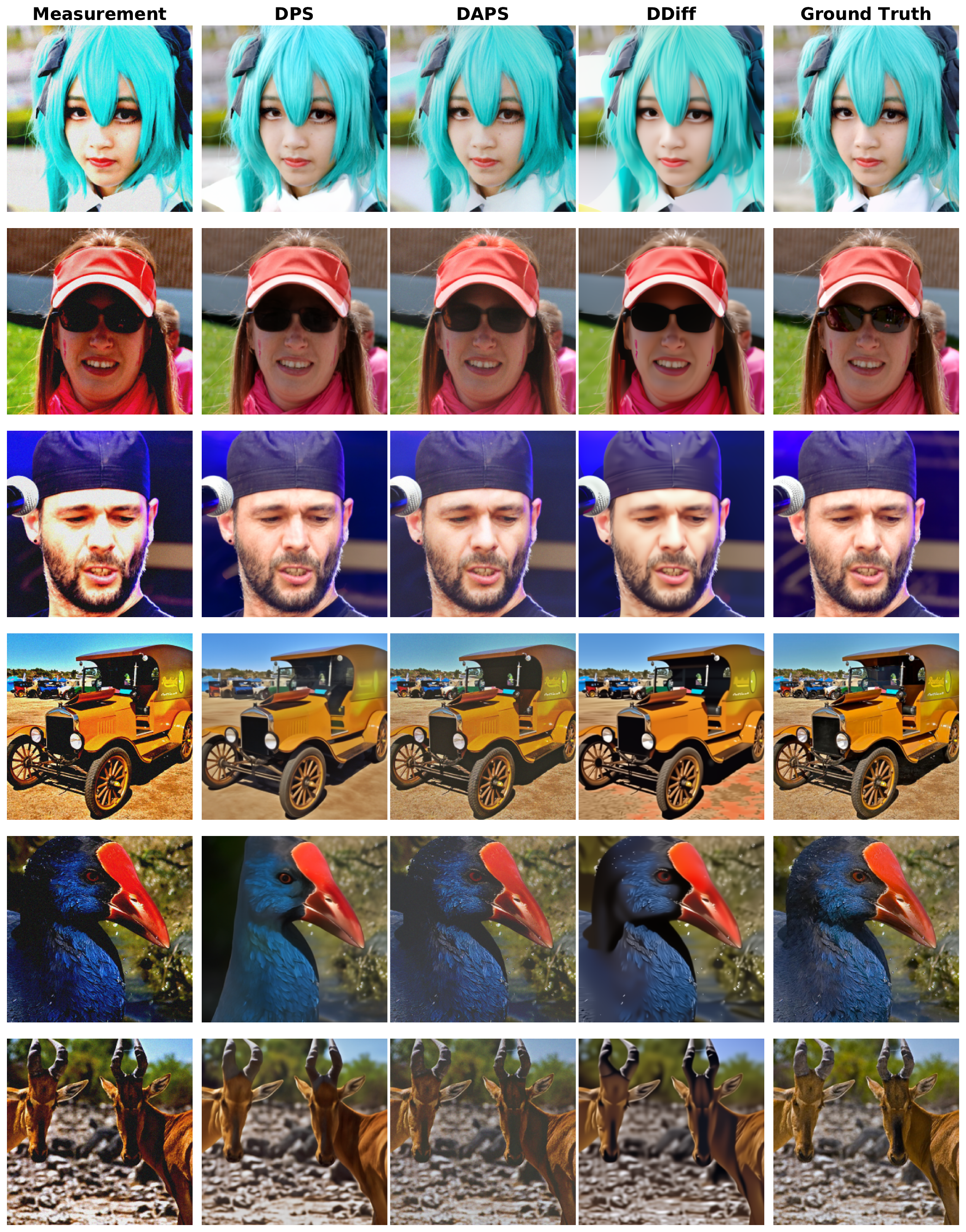}
    \caption{\textbf{Visual comparison of \Method and baselines} on HDR task.}
    \label{fig:hdr_comparison}
\end{figure}

We provide additional qualitative results in Fig.~\ref{fig:superres_comparison} to \ref{fig:hdr_comparison}.

\end{document}